\documentclass{article}

\PassOptionsToPackage{numbers,square,sort&compress}{natbib}
\usepackage[preprint,main]{neurips_2026}

\usepackage[utf8]{inputenc}
\usepackage[T1]{fontenc}
\usepackage{hyperref}
\usepackage{url}
\usepackage{booktabs}
\usepackage{amsfonts}
\usepackage{amsmath}
\usepackage{amssymb}
\usepackage{amsthm}
\usepackage{nicefrac}
\usepackage{microtype}
\usepackage{xcolor}
\usepackage{graphicx}
\usepackage{multirow}
\usepackage{enumitem}
\usepackage{bm}
\usepackage{algorithm}
\usepackage{algpseudocode}
\usepackage{cleveref}
\hypersetup{%
  pdftitle={Practical Non-Stationary Graph Gaussian Processes},
  pdfauthor={K. S. Sesh Kumar, Viacheslav Borovitskiy},
  pdfsubject={Gaussian processes on graphs},
  pdfkeywords={Gaussian processes, graphs, non-stationarity, Matern kernels}}
\usepackage{booktabs}                                                                                               \newcommand{\R}{\mathbb{R}}

\newcommand{\Var}{\mathrm{Var}}
\newcommand{\diag}{\mathrm{diag}}
\DeclareMathOperator*{\argmin}{arg\,min}

\let\textv\v % redefinition: v under letter in text
\renewcommand{\v}{\relax\ifmmode\expandafter\boldsymbol\else\expandafter\textv\fi} % vector
\newcommand{\m}{\expandafter\mathbf} % matrix

\newtheorem{theorem}{Theorem}
\newtheorem{proposition}[theorem]{Proposition}
\newtheorem{definition}{Definition}

\title{Practical Non-Stationary Graph Gaussian Processes}

\author{%
  K.~S.~Sesh Kumar \\
  \normalfont Brevan Howard Centre for Financial Analysis \\
  \normalfont Imperial Business School \\
  \normalfont\texttt{skarri@imperial.ac.uk} \\
  \And
  Viacheslav Borovitskiy \\
  \normalfont School of Informatics \\
  \normalfont University of Edinburgh \\
  \normalfont\texttt{viacheslav.borovitskiy@gmail.com} \\
}

\begin{document}

\maketitle

\begin{abstract}
Mat\'{e}rn Gaussian processes offer a principled framework for learning and uncertainty quantification on graphs. However, they struggle with heterogeneous networks. A central challenge is that variance at a node can represent genuine signal that should strongly \emph{propagate (diffuse)} to neighbors, mere noise that should be \emph{isolated}, or a mixture of both. While this problem has been studied in continuous spatial domains, directly translating those approaches to graphs leaves a free parameter per node, causing extreme over-parameterization. To resolve this, we introduce non-stationary \emph{Variance-Coupled Gaussian Processes (VCGPs)}. These models use an auxiliary signal (often easily derived from historical data) to determine relative diffusivity between nodes, tied to the regression data via a single learned hyperparameter, $\gamma$. The sign of $\gamma$ captures whether the auxiliary signal is directly or inversely proportional to node diffusivity, while its magnitude calibrates this relationship.
We formally prove that $\gamma = 1$ is the unique point where VCGPs are \emph{spectral-stationary} up to rescaling, ensuring that varying $\gamma$ results in genuinely non-stationary correlations. Across five real-world datasets, our VCGPs match or outperform both spectral-stationary and per-node non-stationary baselines.
\end{abstract}

% ==================================================================
\section{Introduction}
\label{sec:intro}
% ==================================================================

Many regression problems involve recovering a latent signal from observations attached to the nodes of a graph. Whether tracking traffic speeds along road networks~\citep{LiYuShahabiBMCYHH2018}, pollutants across urban sensor grids~\citep{LAQN2024}, or temperatures over spatial station networks~\citep{Menne2012}, these applications frequently demand calibrated uncertainty to ensure safety guarantees and drive automated decision-making.
Gaussian processes (GPs) on graphs~\citep{Borovitskiy2021} provide a principled probabilistic framework for this setting. By placing a Gaussian prior over node-indexed functions---as opposed to GPs
whose \emph{inputs} are entire graphs~\citep{Borovitskiy2023}---they encode dependencies through a kernel that reflects the structure of the graph. A common choice is the graph Matérn family~\citep{Whittle1963,Borovitskiy2021,Borovitskiy2020,Mostowsky2025} that extends the widely popular Matérn covariances \cite{porcu2024, RasmussenWilliams2006} to discrete domains via the graph Laplacian.

Despite their appeal, these models impose stationarity and have a single global hyperparameter governing the strength of interactions. Throughout, \emph{stationarity} refers to \emph{spectral stationarity}, defined in Section~\ref{sec:background}; Euclidean translation invariance is not assumed. Real-world networks are rarely so uniform. Traffic congestion concentrates on specific parts of the road network, and pollution dynamics depend on local geography (e.g.\ the altitude profile).
% Capturing such heterogeneity in a principled and computationally tractable manner remains an open challenge.
% For instance, consider a traffic network where each sensor is annotated with a local measure of variability, such as the variance of its readings over the past month.
For instance, consider a road network where some of the nodes have traffic speed readings.
Then the same signal can admit two opposing interpretations. If high variability is due to being an important junction, a node is informative and its fluctuations should \emph{propagate (diffuse)} across the network. If, instead, high variability is due to a malfunctioning sensor, then the node primarily injects noise and its fluctuations should be \emph{isolated}. In practice, which interpretation holds is rarely known a priori; it may vary across datasets and even across regions within a single graph. Designing a model that can express both behaviours in a principled and computationally tractable manner while allowing data to determine the relevant regime is the central question we address.

Existing graph kernels do not provide a simple mechanism to express this distinction. Stationary graph kernels~\citep{SmolaKondor2003,ZhuGhahramaniLafferty2003,Borovitskiy2021} treat all nodes uniformly, ignoring node-specific behaviour. Continuous-domain covariance constructions and variable-coefficient stochastic partial differential equations (SPDEs)~\citep{Gibbs1998,PaciorekSchervish2003,Lindgren2011,Fuglstad2015a,NonstationaryMetricGraphs2025} allow both variance and covariance to vary spatially. These models can use low-dimensional covariate parametrisations; fitting coefficients independently at each vertex instead introduces $O(n)$ parameters (Section~\ref{sec:background}). We introduce an auxiliary-driven non-stationary GP defined directly through the combinatorial graph Laplacian, with an $O(1)$ learned parameter count. It makes non-stationary graph GPs \emph{practical} in this setting by capturing heterogeneous correlations with a constant number of parameters.
% In many systems, variability is not merely marginal noise; it directly modulates how strongly a node influences and responds to its neighbours.

\textbf{Our approach.}
We introduce \emph{Variance-Coupled Gaussian Processes} (VCGPs), a family of graph GPs that captures network heterogeneity using a single interpretable hyperparameter.
To avoid over-parameterization, we leverage an \emph{auxiliary signal} that is frequently available in practice but rarely utilized in kernel design---such as the often-ignored temporal dimension of spatial datasets.
Specifically, from this auxiliary signal, we extract a positive quantity $d_i$ at each node to represent its variability or activity level.
We then define the kernel in terms of both these $d_i$ values and the underlying graph structure.

Importantly, the auxiliary statistics alone do not determine whether high variability should propagate or be isolated. In the traffic example, both a genuinely important junction and a malfunctioning sensor may exhibit large temporal variability and therefore large $d_i$. The distinction instead emerges through the learned coupling parameter $\gamma$. Negative values of $\gamma$ interpret high-variability nodes as informative, amplifying their interactions and allowing their fluctuations to diffuse across the network. Positive values interpret the same variability as unreliable or noisy, suppressing interactions and isolating their influence. The model therefore does not hard-code an interpretation of variability, but instead allows the data itself to determine whether high-$d_i$ nodes are propagators or attenuators of information. When a single global $\gamma$ is insufficient to map variability to diffusivity, VCGPs extend naturally by partitioning the graph into clusters and learning one $\gamma_k$ per cluster. In a road network, for example, high variability may reflect traffic dynamics in one region and sensor faults in another. Cluster-specific exponents allow the model to propagate fluctuations in the former and suppress them in the latter simultaneously, even when their $d_i$ values are similar. This requires regions where variability should propagate and regions where it should be suppressed to lie in different clusters. If an important junction and a faulty sensor have the same $d_i$ and lie in the same cluster, their shared $\gamma_k$ cannot treat their variability differently.

% The sign of $\gamma$ selects the regime (whether node diffusivity is directly or inversely proportional to $d_i$), while its magnitude calibrates the strength of this relationship, with $\gamma \to -\infty$ amplifying diffusivity of high-$d_i$ nodes and $\gamma \to +\infty$ suppressing it. When a single regime is insufficient, the construction extends naturally by partitioning the graph into clusters and learning one $\gamma_k$ per cluster.

\textbf{Contributions.}
\begin{enumerate}[nosep,leftmargin=*]
\item \textbf{Methodology:} We propose Variance-Coupled Gaussian Processes (VCGPs), a novel class of non-stationary graph GPs. By linking node diffusivity to an auxiliary signal, VCGPs can represent complex heterogeneous correlations using a single scalar learnable hyperparameter, $\gamma$.
\item \textbf{Theory:} We prove that $\gamma = 1$ is the unique point where VCGPs share the correlations of a stationary GP while retaining node-dependent variances. It also supplies a warm start (Section~\ref{sec:estimation}): initialising at $\gamma = 1$ retains that fit, so learning $\gamma$ cannot reduce the training marginal likelihood, though held-out negative log-likelihood (NLL) can worsen (PEMS-BAY).
\item \textbf{Evaluation:} Across five real-world datasets, we demonstrate that VCGPs match or outperform both stationary and heavily parameterized non-stationary baselines (\Cref{sec:estimation,sec:experiments}).
\end{enumerate}

\begin{figure}[!t]
\centering
\includegraphics[width=0.88\linewidth]{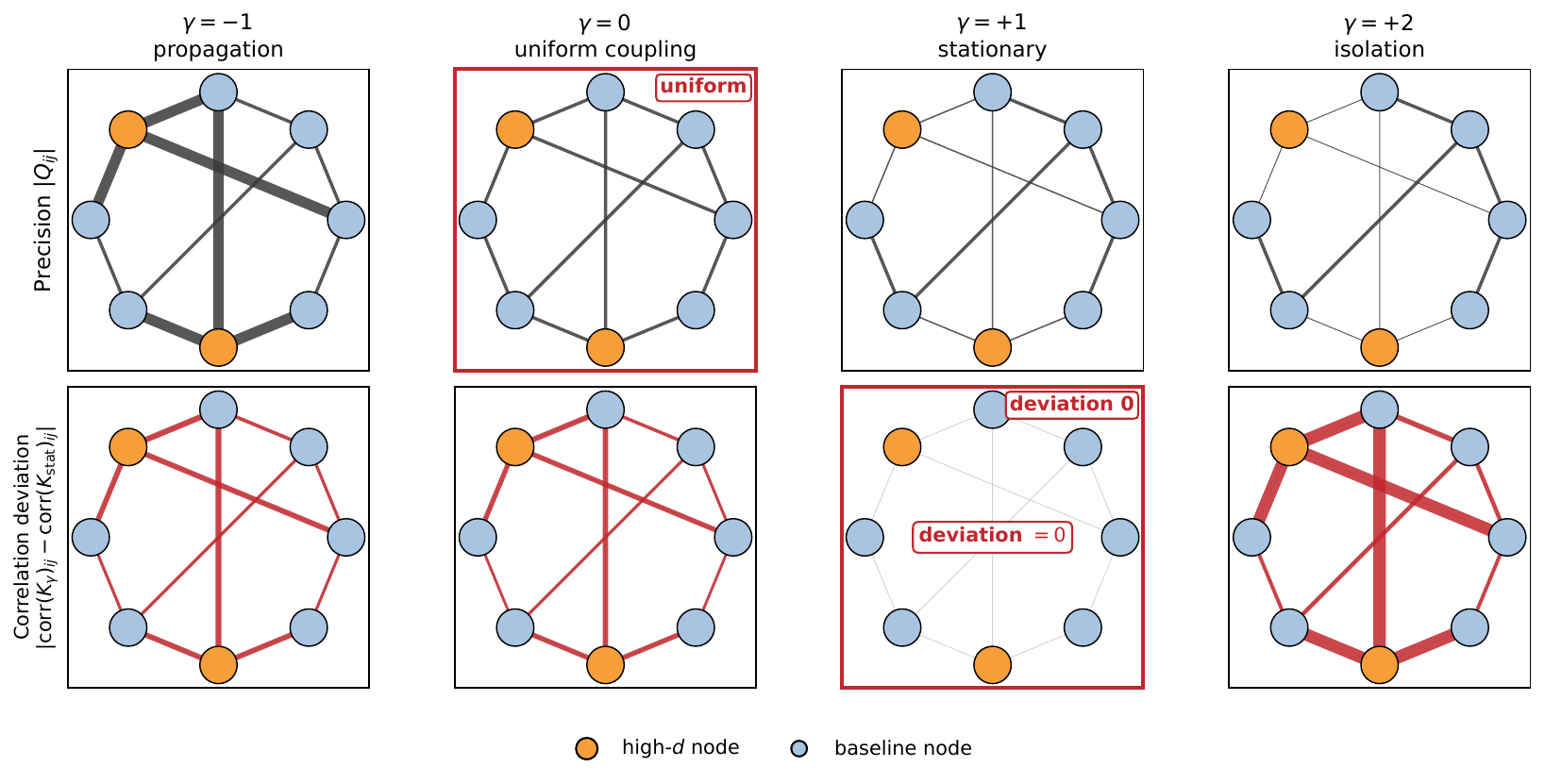}
\caption{\small \textbf{One learned $\gamma$ selects propagation or isolation.}
Orange nodes have high $d_i$; blue nodes are baseline. Edge thickness shows
$|\m{Q}_{ij}|$ (top) and correlation deviation from the stationary baseline (bottom).
For high-$d_i$ nodes, $\gamma<0$ amplifies coupling and $\gamma>0$ suppresses it.
Highlighted: $\gamma=0$, uniform coupling; $\gamma=1$, stationary correlations
with node-dependent variances.}
\label{fig:gamma-schematic}
\vspace{-0.5em}
\end{figure}

% ==================================================================
\section{Background and Motivation}
\label{sec:background}
% ==================================================================

\textbf{Graph Gaussian processes.}
Let $G = (V, E, w)$ be a finite undirected weighted graph with $|V| = n$ nodes, non-negative edge weights $w_{ij} \geq 0$, adjacency matrix $\m{A}_{ij} = w_{ij}$, and (unnormalised) Laplacian $\m{L} = \m{D}_{\mathrm{deg}} - \m{A}$, where $\m{D}_{\mathrm{deg}} = \diag(\sum_j w_{ij})$ collects the weighted node degrees~\citep{Chung1997}. 
 
A graph Gaussian process places a Gaussian distribution over functions $f : V \to \R$, written $f \sim \mathcal{N}(0, \m{K})$, where $\m{K} \in \R^{n \times n}$ is a positive definite kernel matrix. While the kernel is the standard object in probabilistic machine learning, on a graph it is typically dense: an entry $\m{K}_{ij}$ captures the \emph{marginal} co-variation between $f(i)$ and $f(j)$, aggregating contributions from all paths in the graph. For modelling local interactions it is more convenient to work with the precision matrix $\m{Q} = \m{K}^{-1}$~\citep{RueHeld2005}: in a Gaussian Markov random field $\m{Q}_{ij} = 0$ exactly when nodes $i$ and $j$ are conditionally independent given the rest, so the sparsity pattern of $\m{Q}$ encodes the local structure of the graph.

A widely used construction for Gaussian processes on graphs is the graph Mat\'ern family~\citep{Whittle1963,Borovitskiy2021,Borovitskiy2020,Mostowsky2025}, obtained by defining covariance operators as spectral functions of the graph Laplacian; the closely related heat (diffusion) kernel arises in the
$\nu \to \infty$ limit and has proved effective on combinatorial domains
\citep{kondor2002,Doumont2025}. Following the construction of~\citet{Borovitskiy2021}, the graph Mat\'ern process precision operator is defined as
\[
\m{Q}_{\mathrm{Mat\acute ern}}
=
\left(
\frac{2\nu}{\kappa^2}\m{I}
+
\m{L}
\right)^\nu,
\]
where $\kappa > 0$ is a length-scale parameter and $\nu > 0$ controls smoothness. Equivalently, introducing the diffusion parametrisation
\[
2\tau = \frac{\kappa^2}{2\nu},
\qquad\text{so that}\qquad
\frac{1}{2\tau}=\frac{2\nu}{\kappa^2},
\]
the precision can be written, up to a global scaling factor, as
$
\m{Q}_{\mathrm{Mat\acute ern}} 
\propto
\left(
\m{I}
+
2\tau \m{L}
\right)^\nu.
$
Under this parametrisation, $\tau$ directly controls diffusion strength across the graph: larger values propagate information more strongly through neighbouring nodes, while $\nu$ governs spectral decay and hence the regularity of the resulting functions. The simplest case corresponds to $\nu = 1$, yielding
\[
\m{Q}_{\mathrm{stat}}
=
\m{I}
+
2\tau \m{L}.
\]
This serves as our stationary baseline. In all cases, the Laplacian $\m{L}$ induces co-variation between neighbouring nodes, while $\tau$ globally controls interaction strength, so all nodes are treated uniformly.

\textbf{Spectral stationarity.}
Laplacian eigenvectors form the graph Fourier basis, with eigenvalues representing
graph frequencies~\citep{Chung1997,Borovitskiy2021}. Let $\m{U}$ collect these
orthonormal eigenvectors and $\boldsymbol{\Lambda}$ be their diagonal eigenvalue
matrix, so $\m{L}=\m{U}\boldsymbol{\Lambda}\m{U}^{\top}$. For a positive function
$g$ of graph frequency, spectral stationarity means
$\m{K}=g(\m{L})=\m{U}g(\boldsymbol{\Lambda})\m{U}^{\top}$.
Thus the covariance retains the Laplacian's eigenvectors and transforms each
eigenvalue through $g$ to set the corresponding Fourier mode's variance.
The Fourier coefficients are uncorrelated.

\textbf{Stationarity and heterogeneity.}
% Stationarity on graphs corresponds to invariance under graph automorphisms.
While convenient, the stationary graph kernels like graph Mat\'ern may be too restrictive when nodes exhibit different levels of variability or activity. A simple extension introduces heterogeneity through node-specific marginal precisions,
\[
\m{Q}
=
\m{D}^{-1}
+
2\tau \m{L},
\qquad
\m{D}=\diag(d_1,\dots,d_n),
\]
which preserves positive definiteness but only modifies marginal variances, leaving correlations between nodes largely unchanged.

More flexible non-stationary constructions~\citep{Gibbs1998,PaciorekSchervish2003,Lindgren2011,Fuglstad2015,Fuglstad2020} allow both marginal uncertainty and interaction strength to vary across the graph. A direct graph analogue takes the form
\[
\m{Q}
=
\m{D}^{-1}
+
2\tau\,
\m{H}^{1/2}
\m{L}
\m{H}^{1/2},
\qquad
\m{H}=\diag(h_1,\dots,h_n),
\]
where $d_i$ controls marginal variance and $h_i$ controls node-dependent interaction strength. While highly expressive, this construction introduces two free parameters per node, a number of parameters that is impractical to fit from the $\ll n$ observations typically available. %Different combinations of $\{d_i\}$ and $\{h_i\}$ can induce similar covariance structures, making inference unstable and increasing the risk of fitting node-specific fluctuations rather than persistent graph structure.

\textbf{A modeling gap.}
We therefore face a trade-off: stationary models are stable but inflexible, while fully non-stationary models are expressive but difficult to learn. In the next section, we introduce a structured non-stationary family that breaks free from this trade-off by leveraging an auxiliary signal capturing heterogeneity while only using a single additional hyperparameter.

% ==================================================================
\section{Variance-Coupled Gaussian Processes}
\label{sec:family}
% ==================================================================

\textbf{Construction.}
We address the gap from Section~\ref{sec:background} by tying both diagonal and off-diagonal precision structure to a single auxiliary signal $d_i$ at each node $i$. The auxiliary $d_i$ is a fixed, data-derived measure of node variability or activity (e.g.\ temporal variance or coefficient of variation; see Section~\ref{sec:experiments} and Appendix~\ref{sec:appendix-impl-aux}).
We set the marginal precision at node $i$ to $d_i^{-1}$, so a larger $d_i$ corresponds to higher marginal variance. Further, we introduce the scalar parameter $\gamma \in \mathbb{R}$: it determines how variability $d_i$ modulates interaction strength between neighbouring nodes (diffusivity). This separation ensures that $d_i$ encodes marginal variation, while $\gamma$ controls whether (and to what extent) higher variation should lead to propagation or to isolation of information in the node.

%We address the gap from Section~\ref{sec:background} by tying both diagonal and off-diagonal precision structure to a single auxiliary signal $d_i$ at each node, controlled by one scalar $\gamma$. The auxiliary $d_i$ is a fixed, data-derived measure of node variability or activity (e.g.\ temporal variance or coefficient of variation; see Section~\ref{sec:experiments} and Appendix~\ref{sec:appendix-impl-aux}). The diagonal term fixes the marginal precision at node $i$ to $d_i^{-1}$, so that larger $d_i$ corresponds to higher marginal variance. Notably, $\gamma$ does not act on the diagonal: it controls only how variability translates into \emph{interaction strength}. This separation ensures that $d_i$ encodes marginal uncertainty, while $\gamma$ determines whether that variability leads to propagation or isolation across the graph. The coupling between neighbouring nodes $i$ and $j$ is scaled by a factor depending on both $d_i$ and $d_j$, with strength and direction governed by $\gamma$.

\begin{definition}[Variance-coupled precision]\label{def:Q}
Given $\m{D} = \diag(d_1, \dots, d_n)$ with $d_i > 0$ and scalars $\tau, s > 0$ and $\gamma \in \R$, define
\[
  \m{Q}_\gamma \;=\; \m{D}^{-1} + 2\tau\,\m{D}^{-\gamma/2}\,\m{L}\,\m{D}^{-\gamma/2},
  \qquad
  \m{K}_\gamma \;=\; s\,\m{Q}_\gamma^{-1}.
\]
The off-diagonal entries are $\m{Q}_\gamma(i,j) = -2\tau\,d_i^{-\gamma/2} d_j^{-\gamma/2} w_{ij}$, and the $i$-th diagonal entry is $\m{Q}_\gamma(i,i) = d_i^{-1} + 2\tau\,d_i^{-\gamma}\deg(i)$, where $\deg(i) = \sum_j w_{ij}$.
\end{definition}

Only the $d_i^{-1}$ term of the diagonal is $\gamma$-independent: the Laplacian also
contributes $2\tau d_i^{-\gamma}\deg(i)$, and the marginal variance $s\,\m{Q}_\gamma^{-1}(i,i)$
depends on the whole matrix. This is the flexible form of \Cref{sec:background} with the
per-node field pinned to the auxiliary---taking $\m{H} = \m{D}^{-\gamma}$ recovers
$\m{Q}_\gamma$---so its $n$ free entries become one exponent on a measured field, and the
parameter count falls from $O(n)$ to $O(1)$.

The auxiliary signal $\{d_i\}_{i=1}^n$ is fixed, so the learned parameters in Gaussian process regression are the scalar $\gamma$ together with global parameters $(\tau, s, \sigma_{\varepsilon}^2)$, where $\sigma_{\varepsilon}^2$ denotes likelihood (observation) noise.

%\textbf{Interpretation.}
%The parameter $\gamma$ determines how node-level variability translates into interaction strength. Negative values amplify couplings involving high-variability nodes, allowing their influence to propagate across the graph. Positive values suppress these couplings, reducing their influence. Thus, $\gamma$ interpolates between two regimes:
%\begin{itemize}[nosep,leftmargin=*]
%\item \emph{Propagation ($\gamma < 0$):} high-variability nodes strengthen their connections.
%\item \emph{Isolation ($\gamma > 0$):} high-variability nodes weaken their connections.
%\end{itemize}
\textbf{Interpretation.}
The parameter $\gamma$ determines how node-level variability modulates a node’s interactions with its neighbours. For $\gamma < 0$, nodes with larger $d_i$ amplify their connections, increasing their influence on adjacent nodes. For $\gamma > 0$, the same nodes suppress their connections, reducing their influence. Thus, $\gamma$ controls the interactions of a node on the graph.

Figure~\ref{fig:gamma-schematic} illustrates this transition. As $\gamma$ varies, edge strengths and resulting correlations are \emph{amplified} or \emph{suppressed}, with $\gamma = 0$ corresponding to uniform coupling where interaction strength does not depend on $d_i$ and $\gamma = 1$ yielding correlations identical to the stationary baseline of Section~\ref{sec:background} (up to variance rescaling).
We summarise these and some other important regimes below:
% \textbf{Special cases.}
% Several familiar constructions arise as special values of $\gamma$:
\begin{itemize}[nosep,leftmargin=*]

\item $\m{D} = \m{I}, \gamma = 0$:
\[
\m{K}_{\mathrm{stat}}= s\,(\m{I} + 2\tau \m{L})^{-1},
\]
is the stationary kernel as discussed in \Cref{sec:background}; we denote it \texttt{stat}.

\item $\gamma = 0$:
\[
\m{Q}_0 = \m{D}^{-1} + 2\tau \m{L}.
\]
The auxiliary $d_i$ affects only variances, while the interaction strength remains identical to \texttt{stat}. We denote this case \texttt{fem}, after its finite-element form.

%It matches the simplest finite-element discretisation of the variable-coefficient Mat\'ern SPDE~\citep{Lindgren2011}, with $\m{D}^{-1}$ playing the role of the (mass-lumped) mass matrix and $\m{L}$ the stiffness matrix.

\item $\gamma = 1$:
\[
\m{Q}_1 = \m{D}^{-1} + 2\tau \m{D}^{-1/2}\m{L}\m{D}^{-1/2},
\qquad
\m{K}_1 = s\,\m{D}^{1/2}(\m{I}+2\tau\m{L})^{-1}\m{D}^{1/2}.
\]

In this regime, the kernel is obtained from the stationary covariance through a diagonal rescaling. Specifically, if
\[
\m{K}_{\mathrm{stat}} = s\,(\m{I}+2\tau\m{L})^{-1},
\]
then
\[
\m{K}_1 = \m{D}^{1/2}\m{K}_{\mathrm{stat}}\m{D}^{1/2}.
\]
As a result, node-specific variability modifies variances while preserving the underlying correlation structure. Indeed, the covariance between nodes $i$ and $j$ becomes
\[
(\m{K}_1)_{ij}=\sqrt{d_i d_j}\,(\m{K}_{\mathrm{stat}})_{ij}.
\]
% while the variances satisfy
% \[
% (\m{K}_1)_{ii}=d_i\,(\m{K}_{\mathrm{stat}})_{ii}.
% \]
Consequently, the correlation coefficient is unchanged:
\[
\mathrm{Corr}_{\m{K}_1}(i,j) = \frac{(\m{K}_1)_{ij}}{\sqrt{(\m{K}_1)_{ii}(\m{K}_1)_{jj}}}
=
\frac{(\m{K}_{\mathrm{stat}})_{ij}}{\sqrt{(\m{K}_{\mathrm{stat}})_{ii}(\m{K}_{\mathrm{stat}})_{jj}}}.
\]
Thus $\gamma=1$ preserves the same correlation geometry as the stationary model, but allows heterogeneous marginal uncertainty across nodes. We denote this regime by \texttt{het}.
\end{itemize}

\textbf{Amplification and Suppression Regimes.}
The effect of $\gamma$ is explicit in the off-diagonal entries:
\[
\m{Q}_{ij} = -2\tau\,d_i^{-\gamma/2} d_j^{-\gamma/2} w_{ij}.
\]
When $\gamma = 0$, all edges have identical strength. For $\gamma < 0$, edges connected to high-variability nodes are amplified, leading to stronger propagation of influence. For $\gamma > 0$, these edges are suppressed, leading to isolation.

% These changes in precision translate into corresponding changes in correlations. At $\gamma = 1$, the precision factorises as
% \[
% \m{Q}_1 = \m{D}^{-1/2}(\m{I} + 2\tau \m{L}) \m{D}^{-1/2},
% \]
% so the kernel becomes a diagonal scaling of a stationary kernel. This scaling cancels under variance normalisation, and the model behaves identically to the stationary case.

\subsection{Non-stationarity of the VCGP family}
\label{sec:identifiability}

We now show that most values of $\gamma$ yield genuinely non-stationary kernels.

\begin{proposition}[Characterisation of the unique stationary-rescaling point]\label{prop:factorisation}
Let $G = (V, E, w)$ be a connected graph with Laplacian $\m{L}$, and let $d = (d_1,\dots,d_n)$ be non-constant with $d_i > 0$. Define
\[
\m{Q}_\gamma = \m{D}^{-1} + 2\tau\,\m{D}^{-\gamma/2} \m{L} \m{D}^{-\gamma/2},
\qquad
\m{K}_\gamma = s\,\m{Q}_\gamma^{-1},
\]
where $\m{D} = \mathrm{diag}(d)$ and $\tau, s > 0$.
\begin{enumerate}[nosep,leftmargin=*]
\item (\textbf{Stationary rescaling}) At $\gamma = 1$,
\[
\m{K}_1 = s\,\m{D}^{1/2}(\m{I} + 2\tau \m{L})^{-1} \m{D}^{1/2}.
\]
Thus $\m{K}_1$ is a diagonal rescaling of the stationary kernel $\m{K}_{\mathrm{stat}} = s(\m{I} + 2\tau \m{L})^{-1}$ and has the same correlation matrix.
\item (\textbf{Uniqueness}) Assume that any two nodes can be connected through a sequence of node pairs, where each consecutive pair is at distance exactly two, i.e.\ non-adjacent with a common neighbour. Then for any $\gamma \neq 1$, there do not exist a positive diagonal matrix $\m{S}$ and a positive definite matrix $\m{K}_0$ satisfying $\m{K}_0 \m{L} = \m{L} \m{K}_0$ such that $\m{K}_\gamma = \m{S} \m{K}_0 \m{S}$. That is, for $\gamma \neq 1$, $\m{K}_\gamma$ is not a node-wise rescaling of a stationary graph kernel.
\end{enumerate}
\end{proposition}

Proposition~\ref{prop:factorisation} shows that $\gamma=1$ is the unique point at which the auxiliary signal affects only marginal variances without altering the correlation structure. In contrast, for $\gamma \neq 1$, the auxiliary field enters the coupling term $\m{D}^{-\gamma/2} \m{L} \m{D}^{-\gamma/2}$ and cannot be removed by any diagonal rescaling. As a result, the induced correlations are genuinely non-stationary. Variation in $d_i$ makes the kernel non-stationary even with a global $\gamma$, and no member of the family is itself spectral-stationary; a per-node $\gamma_i$ would permit node-level regime changes but reintroduce $O(n)$ parameters, so cluster-specific $\gamma_k$ is the compromise (Section~\ref{sec:extensions}).

This distinction explains the role of $\gamma$: rather than adjusting a redundant scale, it selects between qualitatively different interaction regimes, ranging from propagation to isolation. Away from $\gamma = 1$ the correlations remain well
behaved: $\m{Q}_\gamma$ is a Stieltjes M-matrix, so $\m{K}_\gamma$ is entrywise positive on
a connected graph and every correlation is non-negative at every $\gamma$. The connectivity condition in part (2) is mild: it holds on all of LAQN and on
$98$--$99.5\%$ of vertices for METR-LA, PEMS-BAY and PM2.5, and on the $80\%$ core of CA
Weather's distance-threshold graph, the remainder forming small boundary components
(Appendix~\ref{sec:appendix-coverage}). The proof is given in
Appendix~\ref{sec:appendix-proof}.

\begin{table}[t]
\centering
\caption{Members of the VCGP family. Top: special-value members of \texttt{lgamma}. Bottom: extensions.}
\label{tab:family}
\footnotesize
\setlength{\tabcolsep}{4pt}
\begin{tabular}{@{}llll@{}}
\toprule
Method & Specialisation & Precision $\m{Q}$ / Kernel $\m{K}$ & Free parameters \\
\midrule
\texttt{stat}          & $\m{D} = \m{I}$                          & $\m{Q} = \m{I} + 2\tau \m{L}$                                       & $(\tau, s, \sigma_{\varepsilon}^2)$ \\
\texttt{fem}           & $\gamma = 0$                     & $\m{Q} = \m{D}^{-1} + 2\tau \m{L}$                                   & $(\tau, s, \sigma_{\varepsilon}^2)$ \\
\texttt{het}           & $\gamma = 1$                     & $\m{Q} = \m{D}^{-1} + 2\tau \m{D}^{-1/2} \m{L} \m{D}^{-1/2}$                 & $(\tau, s, \sigma_{\varepsilon}^2)$ \\
\texttt{lgamma}        & learned scalar $\gamma$          & $\m{Q} = \m{D}^{-1} + 2\tau \m{D}^{-\gamma/2} \m{L} \m{D}^{-\gamma/2}$       & $(\tau, s, \sigma_{\varepsilon}^2, \gamma)$ \\
\midrule
\texttt{lgamma\_block} & piecewise $\gamma_k$, $K$ blocks & $\m{Q}_\gamma$ with $\m{L} \to \mathrm{blockdiag}(\m{L}_1, \dots, \m{L}_K)$ & $(\tau, s, \sigma_{\varepsilon}^2, \gamma_1, \dots, \gamma_K)$ \\
\texttt{lgamma\_m15}   & Mat\'ern $\nu = 3/2$             & $\m{K} = s\,\m{Q}_\gamma^{-2}$                                  & same as \texttt{lgamma} \\
\texttt{lgamma\_fisher}& Fisher-info ridge on $\gamma$    & same $\m{Q}$                                                & same as \texttt{lgamma} \\
\bottomrule
\end{tabular}
\end{table}

\subsection{Extensions}
\label{sec:extensions}

The basic scalar-$\gamma$ model admits three direct extensions, summarised in
Table~\ref{tab:family}. First, higher smoothness is obtained by replacing
$\m{K}=s\m{Q}_\gamma^{-1}$ with powers of the same precision, e.g.\
$\m{K}=s\m{Q}_\gamma^{-2}$ for the Mat\'ern-$3/2$ variant
\texttt{lgamma\_m15}. Second, multi-regime graphs are handled by partitioning
the nodes into $K$ clusters and learning one exponent $\gamma_k$ per cluster,
giving \texttt{lgamma\_block}. Third, \texttt{lgamma\_fisher} stabilises estimation by adding a ridge penalty on
$\gamma$, with the regularisation coefficient set from the Fisher information associated
with the marginal likelihood~\cite{Amari1998}.
These modifications compose: \texttt{lgamma\_fisher\_m15} combines the ridge and
Mat\'ern-$3/2$ power. In Table~\ref{tab:leaderboards}, \texttt{block-aug} abbreviates
\texttt{lgamma\_block\_aug}: $K$-means on standardised coordinates augmented with
the standardised auxiliary weighted by $w \geq 0$. The \texttt{aug} suffix
distinguishes this from coordinate-only \texttt{lgamma\_block}.
The label \texttt{spectral-BIC} abbreviates \texttt{spectral\_bic\_lgamma\_block},
which uses spectral clustering with $K$ selected by the training-data
Bayesian Information Criterion (BIC).

The piecewise form used throughout is the \emph{block} form, which also replaces $\m{L}$
by $\mathrm{blockdiag}(\m{L}_1,\dots,\m{L}_K)$ and so makes the clusters conditionally
independent---a modelling assumption. It, the full-edge alternative, the ridge and the
grids over $K$ and $w$ are defined and priced in
Appendix~\ref{sec:appendix-implementation} and Appendix~\ref{sec:appendix-kgrid}.

% ==================================================================
\section{Practical Implementation and Inference}
\label{sec:estimation}
% ==================================================================
We now describe how VCGPs are fit in practice. The goal is to estimate the scalar coupling parameter $\gamma$ together with the global parameters $(\tau, s, \sigma_\epsilon^2)$ from observed data, where $s$ denotes the signal variance (output scale) and $\sigma_\epsilon^2$ the observation noise variance.

\textbf{Marginal likelihood optimisation.}
Given observations $y \in \R^n$ at the graph nodes, we assume the Gaussian observation model
\[
y = f + \epsilon,
\qquad
f \sim \mathcal{N}(0,\m{K}_\gamma),
\qquad
\epsilon \sim \mathcal{N}(0,\sigma_\epsilon^2 \m{I}),
\]
with covariance
\[
\m{K}_\gamma
=
s\,\m{Q}_\gamma^{-1}.
\]
This yields the marginal distribution
\[
y
\sim
\mathcal{N}
\left(
0,\,
\m{K}_\gamma + \sigma_\epsilon^2 \m{I}
\right).
\]

Parameters are estimated by maximising the marginal log-likelihood
\[
\mathcal{L}(\gamma,\tau,s,\sigma_\epsilon^2)
=
-\tfrac12
y^\top
(\m{K}_\gamma+\sigma_\epsilon^2\m{I})^{-1}
y
-\tfrac12
\log\det(\m{K}_\gamma+\sigma_\epsilon^2\m{I})
-\tfrac{n}{2}\log(2\pi).
\]

The optimisation is low-dimensional, with $\gamma$ as the only non-standard parameter. Gradients are computed via standard matrix calculus, and optimisation is performed using off-the-shelf gradient-based methods.

\textbf{Warm-start from the stationary point.}
The model admits a natural reference point at $\gamma = 1$, where the kernel reduces to a variance-rescaled stationary kernel (Section~\ref{sec:identifiability}). This provides a stable and interpretable initialisation. From an Occam's razor perspective, it is the simplest member of the family, before introducing additional non-stationary structure.

In practice, we recommend first fitting the $\gamma = 1$ model and then using the resulting parameters to initialise optimisation of the full model. This warm-start provides a stable starting point, reduces sensitivity to initialization, and allows the optimisation to introduce non-stationarity only when supported by the data.

\textbf{Multi-regime networks.}
When a single value of $\gamma$ is insufficient to capture heterogeneous behaviour, we partition the nodes into $K$ clusters and assign a separate parameter $\gamma_k$ to each cluster.

\looseness=-1
This yields a piecewise VCGP in which coupling adapts locally while retaining a small number of parameters. Clusters are obtained using standard graph clustering methods (e.g.\ spectral clustering), and the parameters $(\gamma_1, \dots, \gamma_K)$ are optimised jointly with the rest of parameters. This extension captures multiple amplification/suppression modulations within a single graph while remaining tractable.

\textbf{Remarks.}
Additional implementation details --- including the auxiliary clipping strategy, the warm-start template, partitioning schemes for the piecewise model, numerical stabilisation (e.g.\ jitter and Cholesky ladder for $\nu = 3/2$), as well as Fisher-information regularisation (\texttt{lgamma\_fisher}) are provided in Appendix~\ref{sec:appendix-implementation}.

\begin{table}[!t]
\centering
\caption{Top-$5$ methods by NLL, plus the strongest outer baseline.
Entries are mean $\pm$ standard error (SE); $^\ast$, $^+$ and $^\wedge$ mark the best NLL, RMSE and
recalibrated NLL among the displayed rows. ``cat'' distinguishes family members
from baselines. \texttt{block-aug} lists $(K,w)$; it and \texttt{spectral-BIC}
abbreviate the corresponding \texttt{lgamma} block members. Auxiliary observations
are disjoint from targets (\Cref{sec:real}).}
\label{tab:leaderboards}
\footnotesize
\setlength{\tabcolsep}{2pt}
\begin{tabular}{@{}l l r r r l@{}}
\toprule
Dataset & method & NLL/pt & RMSE & recNLL/pt & cat \\
\midrule
% ------------------ METR-LA ------------------
\multirow{6}{*}{\shortstack[l]{METR-LA \\ $n_{\mathrm{test}}{=}63$}}
  & \texttt{lgamma\_fisher}                              & $\mathbf{1.2018 \pm 0.0287}^{\ast}$ & $0.8369 \pm 0.0238$ & $1.1763 \pm 0.0232$ & VCGP \\
  & \texttt{lgamma}                                      & $1.2025 \pm 0.0294$ & $0.8468 \pm 0.0247$ & $\mathbf{1.1759 \pm 0.0228}^{\wedge}$ & VCGP \\
  & \texttt{het}                                         & $1.2201 \pm 0.0265$ & $0.8469 \pm 0.0231$ & $1.1975 \pm 0.0234$ & VCGP \\
  & \texttt{lgamma\_fisher\_m15}                         & $1.2202 \pm 0.0344$ & $\mathbf{0.8265 \pm 0.0235}^{+}$ & $1.1816 \pm 0.0247$ & VCGP \\
  & \texttt{lgamma\_m15}                                 & $1.2234 \pm 0.0349$ & $0.8354 \pm 0.0240$ & $1.1835 \pm 0.0244$ & VCGP \\
  & \texttt{fuglreg\_lam1.0}                             & $1.2382 \pm 0.0294$ & $0.8603 \pm 0.0232$ & $1.2064 \pm 0.0224$ & base \\
\midrule
% ------------------ PEMS-BAY ------------------
\multirow{6}{*}{\shortstack[l]{PEMS-BAY \\ $n_{\mathrm{test}}{=}98$}}
  & \texttt{lgamma\_fisher}                              & $\mathbf{1.2720 \pm 0.0346}^{\ast}$ & $0.9015 \pm 0.0360$ & $\mathbf{1.2282 \pm 0.0371}^{\wedge}$ & VCGP \\
  & \texttt{het}                                         & $1.2767 \pm 0.0329$ & $\mathbf{0.8906 \pm 0.0329}^{+}$ & $1.2344 \pm 0.0362$ & VCGP \\
  & \texttt{lgamma}                                      & $1.2835 \pm 0.0370$ & $0.9019 \pm 0.0360$ & $1.2374 \pm 0.0381$ & VCGP \\
  & \texttt{lgamma\_fisher\_m15}                         & $1.2911 \pm 0.0359$ & $0.8963 \pm 0.0329$ & $1.2459 \pm 0.0371$ & VCGP \\
  & \texttt{spectral-BIC}                                & $1.2934 \pm 0.0330$ & $0.9249 \pm 0.0359$ & $1.2582 \pm 0.0329$ & VCGP \\
  & \texttt{fuglreg\_lam1.0}                             & $1.3340 \pm 0.0365$ & $0.8948 \pm 0.0331$ & $1.2918 \pm 0.0349$ & base \\
\midrule
% ------------------ LAQN ------------------
\multirow{6}{*}{\shortstack[l]{LAQN \\ $n_{\mathrm{test}}{=}16$}}
  & \texttt{lgamma\_m15}                                 & $\mathbf{1.4519 \pm 0.0374}^{\ast}$ & $1.0455 \pm 0.0383$ & $1.4068 \pm 0.0332$ & VCGP \\
  & \texttt{lgamma}                                      & $1.4688 \pm 0.0400$ & $1.0434 \pm 0.0393$ & $1.4171 \pm 0.0346$ & VCGP \\
  & \texttt{lgamma\_fisher\_m15}                         & $1.4741 \pm 0.0410$ & $1.0468 \pm 0.0386$ & $1.4227 \pm 0.0355$ & VCGP \\
  & \texttt{lgamma\_fisher}                              & $1.4742 \pm 0.0403$ & $1.0460 \pm 0.0392$ & $1.4223 \pm 0.0347$ & VCGP \\
  & \texttt{spectral-BIC}                                & $1.4779 \pm 0.0528$ & $\mathbf{1.0259 \pm 0.0396}^{+}$ & $\mathbf{1.4020 \pm 0.0372}^{\wedge}$ & VCGP \\
  & \texttt{m05}                                         & $1.5244 \pm 0.0487$ & $1.0447 \pm 0.0397$ & $1.4495 \pm 0.0411$ & base \\
\midrule
% ------------------ PM2.5 ------------------
\multirow{6}{*}{\shortstack[l]{PM2.5 \\ $n_{\mathrm{test}}{=}297$}}
  & \texttt{block-aug} ($20,1$)                          & $\mathbf{0.9756 \pm 0.0106}^{\ast}$ & $\mathbf{0.6541 \pm 0.0055}^{+}$ & $\mathbf{0.9703 \pm 0.0095}^{\wedge}$ & VCGP \\
  & \texttt{block-aug} ($10,1$)                          & $1.0078 \pm 0.0105$ & $0.6822 \pm 0.0075$ & $1.0032 \pm 0.0099$ & VCGP \\
  & \texttt{block-aug} ($10,0.5$)                        & $1.0303 \pm 0.0097$ & $0.7011 \pm 0.0069$ & $1.0269 \pm 0.0090$ & VCGP \\
  & \texttt{block-aug} ($20,0.5$)                        & $1.0410 \pm 0.0121$ & $0.6975 \pm 0.0072$ & $1.0338 \pm 0.0102$ & VCGP \\
  & \texttt{block-aug} ($5,1$)                           & $1.0597 \pm 0.0099$ & $0.7171 \pm 0.0069$ & $1.0553 \pm 0.0095$ & VCGP \\
  & \texttt{fuglreg\_lam1.0}                             & $1.0748 \pm 0.0095$ & $0.7467 \pm 0.0062$ & $1.0705 \pm 0.0082$ & base \\
\midrule
% ------------------ CA Weather ------------------
\multirow{5}{*}{\shortstack[l]{CA Weather \\ $n_{\mathrm{test}}{=}195$ \\ {\scriptsize (mean over $12$ months)}}}
  & \texttt{block-aug} ($10,1.0$)                        & $\mathbf{0.8685 \pm 0.0079}^{\ast}$ & $\mathbf{0.5738 \pm 0.0049}^{+}$ & $\mathbf{0.8572 \pm 0.0076}^{\wedge}$ & VCGP \\
  & \texttt{ps}                                          & $0.9360 \pm 0.0088$ & $0.6261 \pm 0.0053$ & $0.9247 \pm 0.0078$ & base \\
  & \texttt{se\_kriging}                                 & $0.9511 \pm 0.0061$ & $0.6245 \pm 0.0043$ & $0.9376 \pm 0.0065$ & base \\
  & \texttt{spectral-BIC}                                & $0.9562 \pm 0.0229$ & $0.6151 \pm 0.0094$ & $0.9364 \pm 0.0163$ & VCGP \\
  & \texttt{fuglreg\_lam1.0}                             & $1.0004 \pm 0.0087$ & $0.6451 \pm 0.0055$ & $0.9900 \pm 0.0076$ & base \\
\bottomrule
\end{tabular}
\end{table}

% ==================================================================
\section{Experiments}
\label{sec:experiments}
% ==================================================================

In this section we show that VCGPs provide a useful middle ground between stationary graph kernels and heavily parameterised non-stationary models. The experiments ask three questions:
(i) do VCGPs improve prediction and calibration;
(ii) when is the stationary factorisation point $\gamma=1$ sufficient;
and (iii) when is a piecewise multi-regime model necessary?

\textbf{Protocol.}
For each dataset, methods share $20$ $70/30$ train/test splits and the auxiliary.
We report per-point NLL (in nats), root-mean-square error
(RMSE), and recalibrated NLL, which fits one global variance scale on the test
predictions solely as a diagnostic (Appendix~\ref{sec:appendix-impl-baselines}).
Leaderboard settings are chosen post hoc, so these leaderboards do not measure
selection performance; validation-fold selection is assessed separately
(Appendices~\ref{sec:cv}--\ref{sec:aux-ablation}). METR-LA, PEMS-BAY and LAQN seeds
re-split one snapshot rather than giving independent temporal replications.
Auxiliary details are in Appendix~\ref{sec:appendix-impl-aux}.

\textbf{Baselines.}
We compare against six external baselines: \texttt{m05} and \texttt{m15}
(stationary graph Mat\'ern with fixed
$\nu \in \{1/2,3/2\}$~\citep{Borovitskiy2021,Mostowsky2025}),
\texttt{lnu} (graph Mat\'ern with learnable $\nu$), \texttt{ps} (Paciorek--Schervish per-node
lengthscales~\citep{PaciorekSchervish2003,Gibbs1998} sharing our
auxiliary $d_i$ through a single global lengthscale),
\texttt{fuglreg\_lam}$\lambda$ (our graph adaptation of variable-coefficient
SPDE models~\citep{Lindgren2011,Fuglstad2015a}, with per-node $\kappa_i$ and a graph smoothness penalty, the
closest per-node analogue of our one-scalar construction), and
\texttt{se\_kriging}, a squared-exponential-kernel spatial GP fitted by grid search~\citep{RasmussenWilliams2006,Cressie1993}.
Two learned baselines test the family beyond kernel methods: \texttt{dk}, a deep kernel
GP~\citep{WilsonHuNickischSalimbeniDeisenroth2016}, and a five-member deep ensemble of Graph Convolutional
Networks (GCNs)~\citep{KipfWelling2017,Lakshminarayanan2017} taking $d_i$ as a feature. Both lose to the
family leader on raw NLL and RMSE on all five datasets---the ensemble by $0.13$--$3.97$
nat/pt (worst $p = 0.008$) and the deep kernel by $1.11$--$2.89$---though the ensemble is
competitive rather than dominated, its METR-LA deficit being a global variance scale that
recalibration removes (Appendix~\ref{sec:appendix-degnn}). Ensemble calibration on graphs has in any case been
questioned~\citep{Vieira2026}.

\textbf{Datasets.}
We use five real graph regression problems. Throughout, the auxiliary $d_i$ is computed
from observations disjoint from the regression target: a forward window for the traffic
graphs, an interleaved even/odd-day split for PM2.5, a different year for CA Weather, and
an external site-type covariate for LAQN, which uses no temporal signal at all.

\begin{itemize}[nosep,leftmargin=*]
\item \textbf{METR-LA}: Los Angeles freeway sensor graph~\citep{LiYuShahabiBMCYHH2018}.
The signal is the daily mean speed from the first 288 five-minute readings on
2012-03-01. The auxiliary $d_i$ is the coefficient of variation (CV) over the readings
\emph{after} that day---a window disjoint from the target.

\item \textbf{PEMS-BAY}: San Francisco Bay Area freeway sensor graph~\citep{LiYuShahabiBMCYHH2018},
with the same traffic-speed setup and the same disjoint-window
coefficient-of-variation auxiliary.

\item \textbf{LAQN}: London Air Quality Network sensors~\citep{LAQN2024}.
The graph is a 5 km distance-threshold graph. No temporal signal
is used. The values $d_i$ are a six-level site-type ordinal
from rural to kerbside.

\item \textbf{PM2.5}: United States Environmental Protection Agency
Air Quality System annual PM2.5 field for 2023~\citep{EPA_AQS}.
The graph is an 8-nearest-neighbour graph on a set of latitude/longitude locations.
The target is the annual mean over even-ranked days and the auxiliary the standard
deviation (std) over odd-ranked days, so the two share no observation while both stay
representative of the year. We use the standard deviation rather than the coefficient of
variation, preserving level information.

\item \textbf{CA Weather}: Global Historical Climatology Network daily California weather stations~\citep{Menne2012}.
The graph is a 50 km distance-threshold graph. The target is daily maximum temperature
at twelve monthly snapshots in 2023, and the auxiliary is built from 2022, so the variance
field cannot carry target-year information.
\end{itemize}

\textbf{Preliminary results.}
Synthetic recovery experiments confirm that scalar \texttt{lgamma}
recovers $\gamma$ once $|S|/n \geq 0.5$, and a two-regime stochastic
block model shows that \texttt{lgamma\_block} is necessary when
different regions have different propagation/isolation regimes;
see Appendix~\ref{sec:appendix-synthetic} for details.

% ------------------------------------------------------------------
\subsection{Main results}
\label{sec:real}
% ------------------------------------------------------------------

For each dataset we report the best-performing member of the VCGP family and compare it with
the strongest baselines. Table~\ref{tab:leaderboards} consolidates the
top-$5$ leaderboard for every dataset; $^\ast$ marks top NLL, $^+$
marks top RMSE, and $^\wedge$ marks top recalibrated NLL within each
dataset.
Below, we summarize the results.
We defer further discussion of these and some additional results until~\Cref{sec:exp_summary}.

\begin{figure}[!t]
\centering
\includegraphics[width=0.85\linewidth]{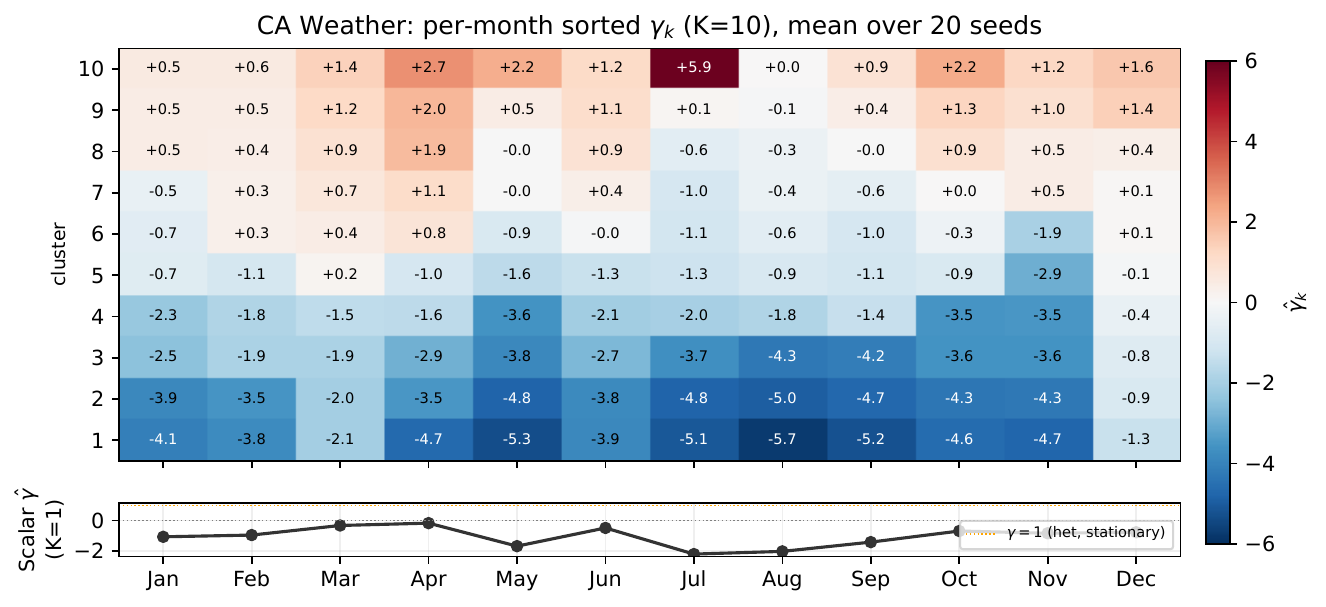}
\caption{CA Weather seasonal heatmap of $\hat\gamma_k$ by month
(\texttt{lgamma\_block}, $K=10$, not the augmented leader; rows are sorted ranks, not
fixed clusters). Most months contain both amplification and suppression regimes, changing
substantially across seasons.}
\label{fig:gamma_seasonal}
\vspace{-0.5em}
\end{figure}
\textbf{METR-LA.}
In METR-LA, the best raw NLL is achieved by \texttt{lgamma\_fisher} ($1.2018$ nat/pt),
while \texttt{lgamma\_fisher\_m15} gives the best
RMSE. This is a useful Pareto split: tuned $\gamma$ improves
uncertainty calibration, while the Mat\'ern-$3/2$ refinement smooths the
mean prediction. The learned exponent is strongly positive over the twenty seeds
($\hat\gamma = +2.9 \pm 0.7$ for scalar \texttt{lgamma}, $+2.1 \pm 0.5$ for the leader),
a strong-isolation regime in which high-variability nodes are down-weighted. Scalar \texttt{lgamma} is statistically indistinguishable from the
Fisher-ridge leader on raw NLL ($1.2025$, $p=0.44$), while every outer
baseline in Table~\ref{tab:metrla_se} is worse at $p<10^{-3}$.

\textbf{PEMS-BAY.}
PEMS-BAY is the clean case of stationary rescaling of Proposition~\ref{prop:factorisation}.
The fixed \texttt{het} model, at the factorisation point $\gamma=1$, gives the best RMSE
($0.8906$) and is statistically indistinguishable from the leading member
\texttt{lgamma\_fisher} on NLL ($1.2767$ against $1.2720$, $p=0.17$); learning $\gamma$
freely is worse ($1.2835$, $p=0.010$) and block variants are worse still. The auxiliary field explains
heterogeneity through variance scaling alone, while the correlation structure stays close to
stationary. 

\textbf{LAQN.}
On LAQN, \texttt{lgamma\_m15} has the best raw NLL ($1.4519$ nat/pt), while \texttt{spectral\_bic\_lgamma\_block} has the best RMSE ($1.0259$) and recalibrated NLL ($1.4020$ nat/pt) in Table~\ref{tab:laqn_se}. BIC selects $K^\star=2$ on 19 of 20 splits; the partition is geographic, not by site type. This block member does not improve raw NLL on scalar \texttt{lgamma} ($+0.01$ nat/pt, $t=+0.4$).

\textbf{PM2.5.}
PM2.5 is the clearest multi-regime case. The leader is
\texttt{lgamma\_block\_aug} with $K=20$ and auxiliary $d=$ annual
standard deviation.
The leader improves over \texttt{m15} substantially on both NLL and RMSE, and considerably outperforms the per-node
SPDE baseline \texttt{fuglreg\_lam1.0}. It does
so with only 20 cluster-wise exponents, compared with 987 per-node
heterogeneity parameters in \texttt{fuglreg}.
The recovered PM2.5 exponents span both regimes, with some clusters in
propagation and others in isolation. This is precisely the setting
where a scalar $\gamma$ is too restrictive and a small vector
$(\gamma_1,\dots,\gamma_K)$ is useful.

\textbf{CA Weather.}
California Weather offers twelve independent monthly snapshots over a fixed graph. The
leader is \texttt{lgamma\_block\_aug} with $(K,w)=(10,1.0)$.
The VCGP gain over its own stationary null is largest on CA Weather, and on the twelve-month
aggregate every outer baseline is beaten at $p<10^{-3}$. Per month, the leader beats all
thirteen outer baselines at $p<0.05$ on both NLL and RMSE in eight of the twelve months
(nine on NLL alone, ten on RMSE alone); requiring Bonferroni correction across the twelve
monthly evaluations leaves five. The winter months are reported as
non-significant headline wins. Figure~\ref{fig:gamma_seasonal} 
shows the per-cluster $\hat\gamma_k$ heatmap across the twelve months.

% ------------------------------------------------------------------
\subsection{Ablation: when does learning $\gamma$ matter?}
\label{sec:ablation}
% ------------------------------------------------------------------

\suppressfloats[b]
\begin{table}[b]
\centering
\caption{Family ablation (NLL/pt). The leader is the lowest-NLL family member in the per-dataset leaderboard. Positive $\Delta$ means improvement: $\Delta_1=\texttt{stat}\!\to\!\texttt{fem}$, $\Delta_2=\texttt{fem}\!\to\!\texttt{het}$, $\Delta_3=\texttt{het}\!\to\!\text{leader}$.}
\label{tab:ablation}
\footnotesize
\setlength{\tabcolsep}{4pt}
\begin{tabular}{@{}llrrrrrrr@{}}
\toprule
Dataset & $d_i$ & \texttt{stat} & \texttt{fem} & \texttt{het} & leader & $\Delta_1$ & $\Delta_2$ & $\Delta_3$ \\
\midrule
METR-LA & CV & 1.2612 & 1.2537 & 1.2201 & 1.2018 & 0.0075 & 0.0336 & 0.0183 \\
PEMS-BAY & CV & 1.3594 & 1.3245 & 1.2767 & 1.2720 & 0.0349 & 0.0478 & 0.0047 \\
LAQN & site-type & 1.6013 & 1.5590 & 1.5234 & 1.4519 & 0.0423 & 0.0355 & 0.0715 \\
PM2.5 & annual std & 1.1329 & 1.1557 & 1.2403 & 0.9756 & -0.0228 & -0.0846 & 0.2647 \\
CA Weather & daily-var & 1.0436 & 1.0673 & 1.1661 & 0.8685 & -0.0237 & -0.0988 & 0.2976 \\
\bottomrule
\end{tabular}
\end{table}

Table~\ref{tab:ablation} compares \texttt{stat}~$\to$~\texttt{fem}~$\to$~\texttt{het}~$\to$~\texttt{leader}.
The steps introduce the auxiliary diagonal $\m{D}$ with uniform coupling,
recover stationary correlations at $\gamma=1$, and test the benefit of learning
a global or cluster-specific $\gamma$.

The ablation separates three regimes. First, PEMS-BAY stops at
\texttt{het}: the auxiliary affects variance, but learning
non-stationary correlations adds no value. Second, METR-LA and LAQN
benefit incrementally from each step, suggesting mild but homogeneous
departures from stationary correlations. Third, PM2.5 and CA Weather
show the opposite pattern: moving to the factorisation point hurts, and
only the cluster-dependent $\gamma$ model recovers the gain. These are
the genuinely multi-regime cases.

The best member is itself diagnostic: \texttt{het} signals stationary correlations after
rescaling, scalar \texttt{lgamma} a homogeneous propagation/isolation regime, and
\texttt{lgamma\_block} spatially varying ones.

\subsection{Experimental summary}
\label{sec:exp_summary}
VCGPs match or outperform stationary and per-node non-stationary baselines,
with the largest gains on multi-regime PM2.5 and CA Weather. PEMS-BAY provides
the complementary case: stationary correlations after variance rescaling suffice.

\textbf{Practical recipe.}
Fit the family in order, warm-starting each extension from its simpler member.
\textbf{(i)~Match the auxiliary to the target's averaging window:}
coefficient of variation for daily snapshots, standard deviation for annual
aggregates, or site type without temporal data. This guidance is based on the
aggregation argument in Appendix~\ref{sec:appendix-impl-aux}, rather than a direct
comparison of coefficient of variation and standard deviation on these datasets.
The PM2.5 validation ablation
compares the external auxiliary with training-target-derived alternatives
(Appendix~\ref{sec:aux-ablation}). A fixed-exponent proxy control illustrates
the risk of a misaligned auxiliary (Appendix~\ref{sec:auxiliary-proxy-control}).
\textbf{(ii)~Always fit \texttt{het} ($\gamma=1$).}
At this unique factorisation point, the auxiliary changes variance while preserving
stationary correlations. It ties the NLL leader and gives the best RMSE on PEMS-BAY.
\textbf{(iii)~Default to \texttt{lgamma\_fisher}:} one learned exponent with a
Fisher-information ridge centred at $\gamma=1$. The ridge adds no parameters and
leads on both traffic graphs, but costs $0.005$ nat/pt on LAQN and $0.018$ on PM2.5.
For mean prediction, the Mat\'ern-$3/2$ refinement replaces $\m{Q}_\gamma^{-1}$ with
$\m{Q}_\gamma^{-2}$ (\texttt{lgamma\_m15}).
\textbf{(iv)~Use blocks when a global exponent is a compromise:}
a spread in fitted $\gamma_k$ or a clear improvement over scalar \texttt{lgamma},
as on PM2.5 and CA Weather. Failure to beat \texttt{het} alone does not justify
escalation: on PEMS-BAY, blocks at $K=2,5,10$ improve test NLL at none of these settings.
\textbf{(v)~Select on validation, not on the leaderboard.}
The primary selection procedure uses a $50/20/30$ split to choose member,
$K$, $d$ and $\lambda$ on validation and score only the selected model on disjoint
test nodes. Appendix~\ref{sec:cv} quantifies its cost relative to post-hoc selection.

%The best family member is itself diagnostic: \texttt{het} signals stationary correlations after rescaling; scalar \texttt{lgamma} signals a homogeneous propagation/isolation regime; \texttt{lgamma\_block} signals spatially varying regimes. Per-baseline formulas, dataset-specific auxiliary choices, the recalibrated-NLL definition, full standard errors, and paired $t$-test $p$-values are collected in Appendix~\ref{sec:appendix-implementation}.

% ==================================================================
\section{Discussion and Conclusion}
\label{sec:discussion}
% ==================================================================

\looseness=-1
We introduced Variance-Coupled Gaussian Processes (VCGPs), a practical family
of non-stationary graph GPs that ties node-level variability to
edge coupling through a single scalar exponent $\gamma$. The construction
keeps the precision sparse, avoids impractical overparameterization,
and has a clear stationary reference point where the kernel is
only a diagonal rescaling of a stationary graph Mat\'ern kernel.

The empirical results show that $\gamma$ is useful not merely as an
extra hyperparameter, but as a diagnostic of graph heterogeneity. On
PEMS-BAY the $\gamma=1$ factorisation point is indistinguishable from the best model and
has the lowest RMSE, indicating stationary correlations after variance rescaling. On METR-LA
and LAQN, learning $\gamma$ gives modest but consistent gains. On PM2.5
and CA Weather, a single global exponent is insufficient; there, a small number
of cluster-wise exponents capture the crucial multi-regime behaviour.

\looseness=-2
\textbf{Limitations.}
The method assumes that a useful auxiliary signal $d_i$ is
available and may be sensitive to its choice.
The present theory establishes the
unique stationary-correlation point and rules out diagonal-rescaling
equivalence away from $\gamma=1$, but does not provide finite-sample
bounds for estimating $\gamma$. Finally, the block extension depends on
a chosen partition family; learning spatially varying $\gamma(x,t)$
directly remains open.

\textbf{Future work.}
Natural extensions include spatio-temporal data and signed Laplacians for
heterophilic graph signals, which can model anti-correlation.

\section*{Acknowledgements and disclosures}

\textbf{Funding and acknowledgements.} K.~S.~Sesh Kumar is funded by the Brevan Howard
Centre for Financial Analysis at Imperial Business School. We thank Dr.~Erik Chavez for providing access to compute
resources.

\textbf{Competing interests.} The authors declare no competing interests. Beyond the
funding stated above, the work generates no income or other benefit to the authors, and
there are no financial relationships outside this work that could be perceived
as influencing it.

\textbf{Reproducibility.} Code is available at
\url{https://github.com/seshkumar/vcgp-graph-gp}. The release documents separate commands for the real-data leaderboards,
validation-fold selection, synthetic studies and the optional Deep Ensemble
comparison. The real-data experiments use the twenty seeds specified in the
protocol; synthetic-study configurations specify their own seeds and settings.
Optional comparisons require the
additional dependencies stated in the release instructions.

% ==================================================================
\bibliographystyle{abbrvnat}
\bibliography{references}

\newpage
\section*{Appendix}
\appendix
% ============================================================
% Appendix — for inclusion at the end of main.tex via:
%
%     \appendix
%     \input{appendix.tex}
%
% Section letters are supplied by the host's \appendix command.
%
% The host main.tex must already provide:
%   \usepackage{booktabs,multirow,enumitem,bm,algorithm,algpseudocode}
%   \newcommand{\m}{\expandafter\mathbf}
%   \newcommand{\diag}{\mathrm{diag}}
%   \newcommand{\R}{\mathbb{R}}
%   \newcommand{\E}{\mathbb{E}}
%   \DeclareMathOperator*{\argmin}{arg\,min}
%   theorem environments (theorem, proposition, lemma, definition, remark)
% ============================================================

This appendix contains: (\textbf{A})~the full proof of
Proposition~1; (\textbf{B})~practical implementation details ---
the fitting algorithm, the per-variant edit cheatsheet, partition
sources for the piecewise extension, the dataset-specific choice
of auxiliary~$d_i$, the baseline tuning protocol, and numerical
detail (M-matrix property, jitter, Cholesky ladder); (\textbf{C})~the
two synthetic experiments (identifiability grid and two-regime
stochastic block model); (\textbf{D})~full per-dataset leaderboards
with standard errors and significance; (\textbf{E})~the deep-ensemble
GNN and deep-kernel comparison; (\textbf{F})~hyperparameters and
numerical settings; (\textbf{G})~the twelve CA Weather monthly
leaderboards; (\textbf{H})~scalability of the kernel factorisation;
(\textbf{I})~long-form heterophily framing and signed-Laplacian
roadmap; (\textbf{J})~per-dataset coverage of the Proposition~1
hypotheses; (\textbf{K})~the $K$-grid for the piecewise variants;
(\textbf{L})~validation-fold selection; (\textbf{M})~choice of
auxiliary and partition source; (\textbf{N})~the auxiliary-proxy control; and
(\textbf{O})~the deep-kernel comparison.

% ==================================================================
\section{Full proof of Proposition~1}
\label{sec:appendix-proof}
% ==================================================================

\paragraph{Statement.}
Let $G = (V, E, w)$ be a connected graph with Laplacian $\m{L}$, and let
$d = (d_1, \ldots, d_n)$ be non-constant with $d_i > 0$. Define
\[
  \m{Q}_\gamma = \m{D}^{-1} + 2\tau\,\m{D}^{-\gamma/2} \m{L} \m{D}^{-\gamma/2},
  \qquad
  \m{K}_\gamma = s\,\m{Q}_\gamma^{-1},
\]
where $\m{D} = \diag(d)$ and $\tau, s > 0$.
\begin{enumerate}[label=(\arabic*),nosep,leftmargin=*]
\item (\emph{Stationary factorisation at $\gamma = 1$}.) At $\gamma = 1$,
  \(
    \m{K}_1 = s\,\m{D}^{1/2}(\m{I} + 2\tau \m{L})^{-1} \m{D}^{1/2}.
  \)
  Hence $\m{K}_1$ is a diagonal rescaling of the stationary kernel
  $\m{K}_{\mathrm{stat}} = s\,(\m{I} + 2\tau \m{L})^{-1}$ and has the same correlation
  matrix.
\item (\emph{Uniqueness}.) Let the \emph{distance-two graph} $D^2(G)$ be the graph on
  the same vertices in which $i \sim j$ exactly when $i \neq j$, $i$ and $j$ are
  \emph{non-adjacent} in $G$, and they share a common neighbour. Assume $D^2(G)$ is
  connected. Then for any $\gamma \neq 1$, there do not exist a positive
  diagonal matrix $\m{S}$ and a positive definite matrix $\m{K}_0$ satisfying
  $\m{K}_0 \m{L} = \m{L} \m{K}_0$ such that $\m{K}_\gamma = \m{S} \m{K}_0 \m{S}$. In particular,
  $\m{K}_\gamma$ is not a node-wise rescaling of any stationary graph kernel
  for $\gamma \neq 1$.
\end{enumerate}

\paragraph{Notation.}
For square matrices $\m{A}, \m{B}$, the commutator is $[\m{A}, \m{B}] = \m{A}\m{B} - \m{B}\m{A}$. Two
symmetric matrices commute if and only if they admit a common
orthonormal eigenbasis. ``$\m{K}_0$ commutes with $\m{L}$'' is the algebraic
form of ``$\m{K}_0$ is diagonalised by the graph Fourier basis'' --- the
standard definition of a stationary graph kernel. For the unnormalised
Laplacian $\m{L} = \m{D}_G - \m{W}$ (where $\m{D}_G$ is the diagonal degree matrix,
distinct from our $\m{D}$), the off-diagonal entries are
$\m{L}_{ij} = -w_{ij} \leq 0$, vanishing exactly when $\{i, j\} \notin E$.

\paragraph{Strategy.}
For $\gamma \neq 1$ the kernel $\m{K}_\gamma$ cannot be written as $\m{S} \m{K}_0 \m{S}$
with $\m{K}_0$ commuting with $\m{L}$. The argument has three steps. (1) The
factorisation hypothesis gives $\m{S} \m{Q}_\gamma \m{S}$ commuting with $\m{L}$;
decomposing into $\Delta + 2\tau\,\Phi \m{L} \Phi$ produces a commutator
identity. (2) Evaluating it on distance-two pairs forces
$\phi_i = \phi_j$, and connectedness of $D^2(G)$ extends this to a
global constant. (3) The leftover diagonal $\Delta$ must then commute
with $\m{L}$; evaluated on edges, this forces $\Delta$ constant on $V$,
which contradicts non-constancy of $d_i^{\gamma - 1}$ when
$\gamma \neq 1$.

\subsection*{Part (1): the factorisation at $\gamma = 1$}

Substituting,
\[
  \m{Q}_1 = \m{D}^{-1} + 2\tau\,\m{D}^{-1/2} \m{L} \m{D}^{-1/2}
      = \m{D}^{-1/2}(\m{I} + 2\tau \m{L}) \m{D}^{-1/2}.
\]
Inverting,
\(
  \m{K}_1 = s\,\m{D}^{1/2} (\m{I} + 2\tau \m{L})^{-1} \m{D}^{1/2}.
\)
Since correlation matrices are invariant under positive diagonal
conjugation, $\m{K}_1$ and $\m{K}_{\mathrm{stat}} = s\,(\m{I} + 2\tau \m{L})^{-1}$ have
identical correlations.

\subsection*{Part (2): uniqueness}

Suppose, for contradiction, that there exist a positive diagonal
$\m{S} = \diag(s_1, \ldots, s_n)$ and a positive definite $\m{K}_0$ with
$[\m{K}_0, \m{L}] = 0$ such that $\m{K}_\gamma = \m{S} \m{K}_0 \m{S}$ for some $\gamma \neq 1$.
From $\m{K}_\gamma = s\,\m{Q}_\gamma^{-1}$ we obtain
$\m{Q}_\gamma = s\,\m{S}^{-1} \m{K}_0^{-1} \m{S}^{-1}$, hence
\[
  M := \m{S} \m{Q}_\gamma \m{S} = s\,\m{K}_0^{-1}.
\]
Since $\m{K}_0^{-1}$ commutes with $\m{L}$, so does $M$.

Define
\[
  \Phi := \m{S} \m{D}^{-\gamma/2} = \diag(\phi_i),
  \qquad
  \Delta := \m{S} \m{D}^{-1} \m{S} = \diag(\Delta_i),
\]
so that $\phi_i = s_i d_i^{-\gamma/2}$ and
$\Delta_i = s_i^2 d_i^{-1} = \phi_i^2 d_i^{\gamma - 1}$. Direct
substitution yields
\begin{equation}\label{eq:M_decomp}
  M = \Delta + 2\tau\,\Phi \m{L} \Phi.
\end{equation}
The condition $[M, \m{L}] = 0$ then gives
\begin{equation}\label{eq:commutator}
  [\Delta, \m{L}] = -2\tau\,[\Phi \m{L} \Phi, \m{L}].
\end{equation}

\paragraph{Step 1: $\phi$ is constant on distance-two pairs.}
Let $i, j$ be non-adjacent at graph distance two. Then $\m{L}_{ij} = 0$, so
the $(i, j)$ entry of $[\Delta, \m{L}]$ vanishes. A direct computation,
using the diagonal-matrix identities $(\Phi \m{A})_{ij} = \phi_i \m{A}_{ij}$
and $(\m{A} \Phi)_{ij} = \m{A}_{ij} \phi_j$, gives
\[
  [\Phi \m{L} \Phi, \m{L}]_{ij} = (\phi_i - \phi_j)\,(\m{L} \Phi \m{L})_{ij}.
\]
Here $(\m{L} \Phi \m{L})_{ij} = \sum_k \m{L}_{ik}\,\phi_k\,\m{L}_{kj}$, with non-zero
contributions only from common neighbours $k \in N(i) \cap N(j)$, where
$\m{L}_{ik}, \m{L}_{kj} < 0$ and $\phi_k > 0$. Each such term is strictly
positive, and at least one common neighbour exists by hypothesis, so
$(\m{L} \Phi \m{L})_{ij} > 0$. Hence $\phi_i = \phi_j$ for every distance-two
pair, and by connectedness of $D^2(G)$, $\phi_i = c$ for all $i$, with
some $c > 0$.

\paragraph{Step 2: $\Delta$ cannot commute with $\m{L}$.}
With $\phi \equiv c$, $\Phi \m{L} \Phi = c^2 \m{L}$, so
$M = \Delta + 2\tau c^2 \m{L}$. The condition $[M, \m{L}] = 0$ reduces to
$[\Delta, \m{L}] = 0$. For any edge $(i, j) \in E$,
\[
  [\Delta, \m{L}]_{ij} = (\Delta_i - \Delta_j)\,\m{L}_{ij} = 0,
\]
and $\m{L}_{ij} \neq 0$ on edges, so $\Delta_i = \Delta_j$ for every edge.
By connectedness of $G$, $\Delta_i$ is constant over $V$.

But $\Delta_i = c^2 d_i^{\gamma - 1}$, which is non-constant whenever
$d$ is non-constant and $\gamma \neq 1$. This is a contradiction.
Therefore no such factorisation exists, and $\m{K}_\gamma$ is not a
node-wise rescaling of any stationary graph kernel for
$\gamma \neq 1$.\hfill$\square$

\paragraph{Algebraic remark on $\gamma = 0$.}
At $\gamma = 0$, the precision $\m{Q}_0 = \m{D}^{-1} + 2\tau \m{L}$ has
off-diagonals $-2\tau w_{ij}$ that do not depend on $d$, while the
diagonal carries the auxiliary as $\m{D}^{-1}$. This is the
mass-lumped finite-element discretisation of a variable-coefficient
Mat\'ern SPDE with $\m{D}^{-1}$ playing the role of the (mass-lumped)
mass matrix and $\m{L}$ the stiffness matrix~\citep{Lindgren2011};
we call this kernel \texttt{fem} in the family.
The further specialisation $\m{D} = \m{I}$ recovers the semi-supervised graph
kernel $\m{Q} = \m{I} + 2\tau \m{L}$ of~\citep{ZhuGhahramaniLafferty2003,
SmolaKondor2003}, which we call \texttt{stat}.

The identity at $\gamma = 0$ does \emph{not} assert that $\m{K}_0$ is
stationary: when $d$ is non-constant, $\m{D}^{-1}$ does not commute with
$\m{L}$, so $\m{Q}_0$ does not either, and $\m{K}_0$ is genuinely non-stationary
on the graph. The uniqueness argument above applies verbatim at
$\gamma = 0$ (with $d_i^{1 - \gamma} = d_i$ non-constant), confirming
that \texttt{fem} is not a node-wise rescaling of any stationary graph
kernel.

% ==================================================================
\section{Practical implementation: full details}
\label{sec:appendix-implementation}
% ==================================================================

This section gives the implementation details deferred from~\S4 of the main paper: the fitting algorithm and warm-start template (\S\ref{sec:appendix-impl-algo}), the per-variant edit cheatsheet (\S\ref{sec:appendix-impl-variants}), partition sources for the piecewise extension (\S\ref{sec:appendix-impl-clusters}), the dataset-specific choice of auxiliary~$d_i$ (\S\ref{sec:appendix-impl-aux}), the baseline tuning protocol (\S\ref{sec:appendix-impl-baselines}), and numerical detail (\S\ref{sec:appendix-impl-numerical}).

\subsection{Algorithm and warm-start}
\label{sec:appendix-impl-algo}

\paragraph{Marginal log-likelihood.}
A VCGP is fitted by joint maximisation of the marginal log-likelihood (MLL) over $(\log\tau, \log s, \log\sigma^2, \gamma)$ on a training subset $\m{S} \subset V$ with targets $y_\m{S}$. With the Gaussian observation model $y_\m{S} \sim \mathcal{N}(0,\,\m{K}_{\gamma, SS} + \sigma^2 \m{I})$,
\[
\mathcal{L}(\theta) \;=\; -\tfrac{1}{2}\,y_\m{S}^\top (\m{K}_{\gamma, SS} + \sigma^2 \m{I})^{-1} y_\m{S}
                         \;-\; \tfrac{1}{2}\,\log\det(\m{K}_{\gamma, SS} + \sigma^2 \m{I})
                         \;-\; \tfrac{|\m{S}|}{2}\,\log(2\pi).
\]
We optimise with Adam for $200$--$300$ iterations. Gradients flow through $\m{Q}_\gamma$ via standard automatic differentiation; the only non-standard term is the log-determinant of $\m{K}_{\gamma, SS} + \sigma^2 \m{I}$, computed via Cholesky.

\paragraph{Stationary warm-start.}
At $\gamma = 1$, Proposition~1 makes the VCGP equivalent to a stationary kernel rescaled by $\sqrt{d_i}$ on the diagonal. We exploit this for a principled initialisation: first fit the three-parameter \texttt{het} ($\gamma$ frozen at $1$), then add $\gamma$ as a free fourth scalar, initialising it at~$1$. The MLL is therefore non-decreasing along the warm-start --- the \texttt{lgamma} fit can only improve on \texttt{het}, by construction. We initialise each cluster $\gamma_k$ at~$1$ in the piecewise variant \texttt{lgamma\_block}.

\begin{algorithm}[h]
\caption{Scalar-$\gamma$ VCGP fit.}
\label{alg:fit}
\begin{algorithmic}[1]
\Require Laplacian $\m{L}$, auxiliary $d \in \R^n_{>0}$, training set $\m{S} \subset V$ with targets $y_\m{S}$, iterations $T$.
\Ensure Posterior mean and variance $(\mu_*, \sigma^2_*)$ at every node.
\State $d_i \gets \mathrm{clip}(d_i/\bar d,\,0.1,\,10)$; \quad $\m{D} \gets \diag(d)$
\State Warm-start: fit \texttt{het} ($\gamma = 1$); initialise $(\log\tau, \log s, \log\sigma^2, \gamma)$ with $(\log\tau_h, \log s_h, \log\sigma^2_h,\,1)$
\For{$t = 1, \dots, T$}
  \State $\m{Q}_\gamma \gets \m{D}^{-1} + 2\tau\,\m{D}^{-\gamma/2}\,\m{L}\,\m{D}^{-\gamma/2}$
  \State $\m{K}_\gamma \gets s\,(\m{Q}_\gamma + \varepsilon \m{I})^{-1}$
  \State Adam step on $-\log p(y_\m{S} \mid \theta)$, $\theta = (\log\tau, \log s, \log\sigma^2, \gamma)$
\EndFor
\State $\mu_* \gets \m{K}_{\cdot \m{S}}(\m{K}_{SS}+\sigma^2 \m{I})^{-1} y_\m{S}$;\quad
       $\sigma^2_*(i) \gets \m{K}_{ii}+\sigma^2 - \m{K}_{iS}(\m{K}_{SS}+\sigma^2 \m{I})^{-1}\m{K}_{Si}$
\end{algorithmic}
\end{algorithm}

\subsection{Per-variant edits}
\label{sec:appendix-impl-variants}

Every learned-$\gamma$ variant is a single-line edit of Algorithm~\ref{alg:fit} (Table~\ref{tab:variants}). The two refinements commute and compose: \texttt{lgamma\_fisher\_m15} applies both \texttt{lgamma\_m15} and \texttt{lgamma\_fisher}.

\begin{table}[h]
\centering
\caption{Each variant is a single-line edit of Algorithm~\ref{alg:fit}.}
\label{tab:variants}
\footnotesize
\setlength{\tabcolsep}{4pt}
\begin{tabular}{@{}l l l@{}}
\toprule
Member & Line & Edit \\
\midrule
\texttt{lgamma\_block}   & 4 & replace $\m{L}$ with $\m{L}_{\mathrm{bl}} = \mathrm{blockdiag}(\m{L}_1, \dots, \m{L}_K)$; learn $\gamma \in \R^K$ \\
\texttt{lgamma\_m15}     & 5 & $\m{K}_\gamma \gets s\,(\m{Q}_\gamma + \varepsilon \m{I})^{-2}$ \\
\texttt{lgamma\_fisher}  & 6 & add $\lambda(\gamma - \gamma_{\mathrm{ref}})^2$ to the loss
                                ($\lambda = 0.5/\Var(\hat\gamma_{\mathrm{vanilla}})$, $\gamma_{\mathrm{ref}} = 1$) \\
\bottomrule
\end{tabular}
\end{table}

\paragraph{Smoother kernels (\texttt{lgamma\_m15}).}
Replacing $\m{K}_\gamma = s\,\m{Q}_\gamma^{-1}$ with $\m{K}_\gamma = s\,\m{Q}_\gamma^{-2}$ gives the Mat\'ern-$3/2$ member of the family. This adds no new parameters and is implementable as two back-solves through $\m{Q}_\gamma$.

\textbf{What the Fisher-information ridge is.} The quantity used is an entry of the full
inverse Hessian. We first fit the plain model by marginal likelihood over
$(\tau, s, \sigma_\varepsilon^2, \gamma)$; at the estimate we form the $4\times4$ Hessian
$\m{J}$ of the negative log marginal likelihood, invert it, and take
$\Var(\hat\gamma) = \m{J}^{-1}(\gamma,\gamma)$, and add a quadratic penalty pulling $\gamma$
toward the factorisation point with weight $\lambda = 0.5/\Var(\hat\gamma)$. To quadratic
order this pulls $\hat\gamma$ exactly halfway to $\gamma = 1$: a scale-free shrinkage that
needs no tuning.

\paragraph{Stabilisation on small datasets (\texttt{lgamma\_fisher}).}
When $|\m{S}|$ is small, the MLL surface for $\gamma$ can be flat. We add a quadratic ridge $\lambda(\gamma - 1)^2$ anchored at the stationary point, with $\lambda$ set automatically from the variance of the unregularised $\hat\gamma$ on the same training set. The penalty is read once off the inverse Hessian of the vanilla fit; no manual tuning is required.

\paragraph{What the block model assumes.}
The \emph{full-edge} form keeps every edge and varies
only the exponent by region, one $\gamma_k$ per cluster. The \emph{block} form,
\texttt{lgamma\_block}, additionally replaces $\m{L}$ by
$\mathrm{blockdiag}(\m{L}_1,\dots,\m{L}_K)$, removing cross-cluster edges and so making
the clusters conditionally independent. This is a modelling assumption and we state it as
such; it costs little where regions are genuinely near-independent, as with CA Weather's
distinct climate zones. Whether it pays is empirical: on PM2.5, keeping every edge costs
$+0.044 \pm 0.007$ negative log-likelihood per test point (NLL/pt) against the block
variant. That comparison comes from the validation-fold runs of
\S\ref{sec:cv} (full-edge $0.9918$ against block $0.9482$ on the held-out fold), not from
the $70/30$ protocol used elsewhere in this section, because the full-edge variant was only
ever run inside the selection grid.

\subsection{Partition sources for the piecewise extension}
\label{sec:appendix-impl-clusters}

For a fixed cluster count $K$, the partition is computed before fitting the
per-cluster exponents. We use three partition rules:
\begin{itemize}[nosep,leftmargin=*]
\item \textsc{plain} $K$-means on node coordinates $[\mathrm{lat}_i, \mathrm{lon}_i]$ (\texttt{lgamma\_block});
\item \textsc{augmented} $K$-means on $[\widetilde{\mathrm{lat}}_i, \widetilde{\mathrm{lon}}_i,\, w \cdot \tilde d_i]$, where tildes denote standardisation of each coordinate and the auxiliary, and $w \geq 0$ controls the auxiliary's influence (\texttt{lgamma\_block\_aug}, labelled \texttt{block-aug} in Table~\ref{tab:leaderboards});
\item \textsc{spectral} $K$-means on the first $K$ eigenvectors of the unnormalised graph Laplacian, with $K$ minimising the training-data criterion $-2\log p(y_\m{S} \mid \hat\theta_K) + K\log|\m{S}|$ over the fitted grid (\texttt{spectral\_bic\_lgamma\_block}, labelled \texttt{spectral-BIC} in Table~\ref{tab:leaderboards}).
\end{itemize}
All three feed the same \texttt{lgamma} fit, with per-cluster $\gamma_k$ initialised at~$1$.
For each partition, the implementation rebuilds a symmetric geographic
nearest-neighbour graph within each cluster, using coordinates alone for edge
construction and excluding cross-cluster edges. Thus the block Laplacian is
formed from these within-cluster graphs, rather than merely deleting edges
from the original graph. The auxiliary affects the augmented partition and
the precision, but does not enter the within-cluster distance calculation.

\subsection{Variance field $d_i$}
\label{sec:appendix-impl-aux}

The choice of $d_i$ is dictated by the averaging window of the predicted signal, not by what is convenient to compute.

\textbf{(i) Daily snapshots with similar baselines} (METR-LA, PEMS-BAY, LAQN). Temporal coefficient of variation $d_i = \mathrm{std}/\mathrm{mean}$ across the source readings, or, on LAQN (no temporal series), a six-level site-type ordinal $\{\textsc{Rural}\!\to\!0, \ldots, \textsc{Kerbside}\!\to\!5\}$, normalised to mean $1$ and then shifted by $+0.1$ before the clip, so that the lowest category is strictly positive as Definition~1 requires (only five of the six categories occur in the cached data).

\textbf{(ii) Aggregates over longer windows} (PM2.5 annual mean from $\sim\!365$ daily reads). The aggregate's residual variance scales with daily-std$^2/N$, not CV$^2$, and dividing by the mean discards level information; we use $d_i =$ annual std on PM2.5 and $2022$ daily-mean variance on CA Weather.

The choice of $d_i$ for each dataset is post-hoc. On PM2.5,
Table~\ref{tab:pm25-aux} compares the external standard-deviation
auxiliary with local variability estimated from training targets; it does
not compare standard deviation with coefficient of variation.

\subsection{Baseline tuning protocol}
\label{sec:appendix-impl-baselines}

\paragraph{Per-baseline formulas and parameter counts.}
\texttt{m05}, \texttt{m15}, \texttt{lnu}: spectral graph Mat\'ern kernels at $\nu = 1/2$, $\nu = 3/2$, and learnable $\nu$ respectively, parametrised by lengthscale $\ell$ and amplitude; $3$ parameters (\texttt{m05}, \texttt{m15}) or $4$ (\texttt{lnu}). \texttt{ps}: Paciorek--Schervish kernel with per-node lengthscale $\ell_i = \ell_0 / \sqrt{d_i}$ tied through one global $\ell_0$, $3$ parameters. \texttt{fuglreg\_lam}$\lambda$: our graph adaptation of variable-coefficient Mat\'ern SPDE models~\citep{Lindgren2011,Fuglstad2015a}, with per-node $\kappa_i$ and graph smoothness penalty $\lambda \sum_{(i,j) \in E} (\log \kappa_i - \log \kappa_j)^2$, $n + 3$ parameters; the closest per-node analogue of our one-scalar construction. \texttt{se\_kriging}: a squared-exponential-kernel spatial GP~\citep{RasmussenWilliams2006} fitted by grid search over lengthscale, amplitude and noise; $3$ parameters, and the standard kriging predictor of classical spatial statistics~\citep{Cressie1993}. This is the spatial-kriging baseline reported on all five datasets. An implementation of Bayesian kernelised tensor regression~\citep{LeiLabbeSun2025} ships with the code for reference, but its output is not the row reported here.

\paragraph{Fitting protocol.}
Stationary Mat\'ern baselines (\texttt{m05}, \texttt{m15}, \texttt{lnu}) share the family's Adam-MLL protocol; \texttt{ps}, \texttt{se\_kriging}, and \texttt{fuglreg\_lam}$\lambda$ are fitted as published. We sweep $\lambda \in \{10^{-2}, 10^{-1}, 1, 10, 10^2\}$ per dataset and report the post-hoc best as \texttt{fuglreg\_lam1.0}; this is fixed-grid, not cross-validated per seed, and the neighbouring $\lambda$ values are substantially worse.

\paragraph{Recalibrated NLL/pt.}
The recalibrated NLL fits a single test-side scalar
\[
c^* = \argmin_c \mathrm{NLL}(c\hat\sigma^2)
\]
that rescales predictive variances. This is reported as a calibration
diagnostic, not as a deployable correction --- a deployable corrector
would have to be fit on a held-out validation set, not on the test set.

\paragraph{The two remaining outer baselines.}
\texttt{specpoly} is a spectral-polynomial kernel, $\m{K} = g(\m{L})$ with $g$ a learned
degree-$3$ polynomial in the Laplacian spectrum --- the most flexible
\emph{spectral-stationary} comparator, and so a direct test of whether the gains come
from non-stationarity rather than from a richer spectral filter.
\texttt{fuglstad} is the variable-coefficient SPDE with per-node $\kappa_i$ and
\emph{no} smoothness penalty, i.e.\ \texttt{fuglreg\_lam}$\lambda$ at
$\lambda = 0$; it is reported to show what the per-node construction does when its
$O(n)$ parameters are left unregularised.

\paragraph{Significance testing.}
Per-row paired one-sided $t$-tests against the dataset leader are computed across the $20$ seeds, with all methods evaluated on the same train/test splits to control for split-level variance. The leaderboards and the CA Weather monthly analysis use one-sided tests throughout; the two-method comparisons quoted in \S\ref{sec:appendix-degnn} are two-sided, which is the more conservative choice when neither direction is privileged, and are labelled as such there. Standard errors in the leaderboards are seed-level SEs ($s/\sqrt{20}$).

\paragraph{Two caveats on reading the leaderboards.}
First, on the three small single-snapshot graphs the twenty seeds are re-splits of
\emph{one} snapshot, not independent replications, and the test sets are small
($n_{\mathrm{test}} \approx 16$ on LAQN, $63$ on METR-LA). We therefore rest the
multiplicity-corrected claim on CA Weather's twelve monthly snapshots and read the
single-snapshot datasets as corroborating rather than carrying the result.
Second, both sides of the leaderboard comparison use post-hoc-best settings: the
baselines keep their best sweep value --- \texttt{fuglreg\_lam1.0} is the best of a
five-point $\lambda$ sweep --- and the reported family member and per-dataset $d_i$ are
likewise chosen post hoc. The comparison is therefore not conservative in either
direction. We also run the selection mechanically on a disjoint validation fold
(\S\ref{sec:cv}, \S\ref{sec:aux-ablation}); that honest protocol is worse than reading
the leaderboard on every dataset, by $0.04$ nat/pt on PM2.5 and $0.07$ on CA Weather,
rising to $0.23$--$0.28$ on the three small graphs, where it also fits on a smaller
$50/20/30$ split.

\subsection{Numerical detail}
\label{sec:appendix-impl-numerical}

\paragraph{M-matrix property.}
$\m{Q}_\gamma = \m{D}^{-1} + 2\tau\,\m{D}^{-\gamma/2} \m{L} \m{D}^{-\gamma/2}$ has positive diagonal $\m{D}^{-1}_{ii} + 2\tau d_i^{-\gamma} \m{D}_{ii}^{\mathrm{deg}}$ and non-positive off-diagonal $-2\tau d_i^{-\gamma/2} d_j^{-\gamma/2} w_{ij}$ ($w_{ij} \geq 0$ on edges, zero off-edges). It is therefore a Stieltjes M-matrix: positive definite, with non-negative inverse $\m{K}_\gamma$. Entrywise positivity is preserved under the block-diagonal generalisation $\m{L}_{\mathrm{bl}}$.

\paragraph{$\m{Q}_\gamma$ inversion jitter.}
Direct inversion of $\m{Q}_\gamma$ uses sparse LU (\texttt{scipy.sparse.linalg.splu}) when $\mathrm{nnz}(\m{Q}_\gamma) < n^2/4$ and dense \texttt{linalg.inv} otherwise. A small jitter $10^{-8}\,\m{I}$ is added to the diagonal before inversion.

\paragraph{Cholesky jitter ladder.}
For the Mat\'ern-$3/2$ refinement $K = s\,\m{Q}_\gamma^{-2}$, the squared precision raises the condition number; the Cholesky of $\m{K}_{SS} + \sigma^2 \m{I}$ uses an escalating jitter sequence to recover from rare \texttt{LinAlgError} events: $\{10^{-8}, 10^{-7}, \ldots, 10^{-2}\}$ in the shared kernel code and the shorter $\{10^{-6}, \ldots, 10^{-2}\}$ on the $\nu = 3/2$ path. The ladder fires more often at $\nu = 3/2$ than at $\nu = 1/2$, as the squared precision would suggest, and in no case does it persist past a single Adam iteration.

\paragraph{Stability under $d_i$ rescaling.}
Multiplying $d_i$ by a global constant rescales $\m{D}^{-1}$ and $\m{D}^{-\gamma/2}$ jointly; the construction is invariant up to a redefinition of $\tau$ and $s$. We normalise $d_i$ to mean~$1$ before fitting.

% ==================================================================
\section{Synthetic experiments}
\label{sec:appendix-synthetic}
% ==================================================================

Two controlled synthetic studies accompany the real-data analysis: a
single-regime grid that verifies scalar $\gamma$ recovery (a
prerequisite for any real-data $\hat\gamma$ to be meaningful) and a
two-regime stochastic block model that tests the piecewise extension
on a multi-regime ground truth.

\subsection{Single-regime identifiability of scalar $\gamma$}
\label{sec:appendix-synth-single}

\paragraph{Setup.}
A $20 \times 20$ regular grid ($n = 400$); $\m{D} = \diag(d_i)$ varies
linearly from $0.2$ to $2.0$ along the horizontal coordinate so the
heterogeneity needed to identify $\gamma$ is present. For each ground
truth $\gamma_{\mathrm{true}} \in \{-1, 0, +1, +3\}$ and each training
fraction $|\m{S}|/n \in \{0.1, 0.2, 0.3, 0.5, 0.7, 0.9\}$ we draw $10$ seeds and
fit scalar \texttt{lgamma} by joint MLL (Adam, $300$ steps,
warm-started at $\gamma = 1$).

\begin{figure}[h]
\centering
\includegraphics[width=\textwidth]{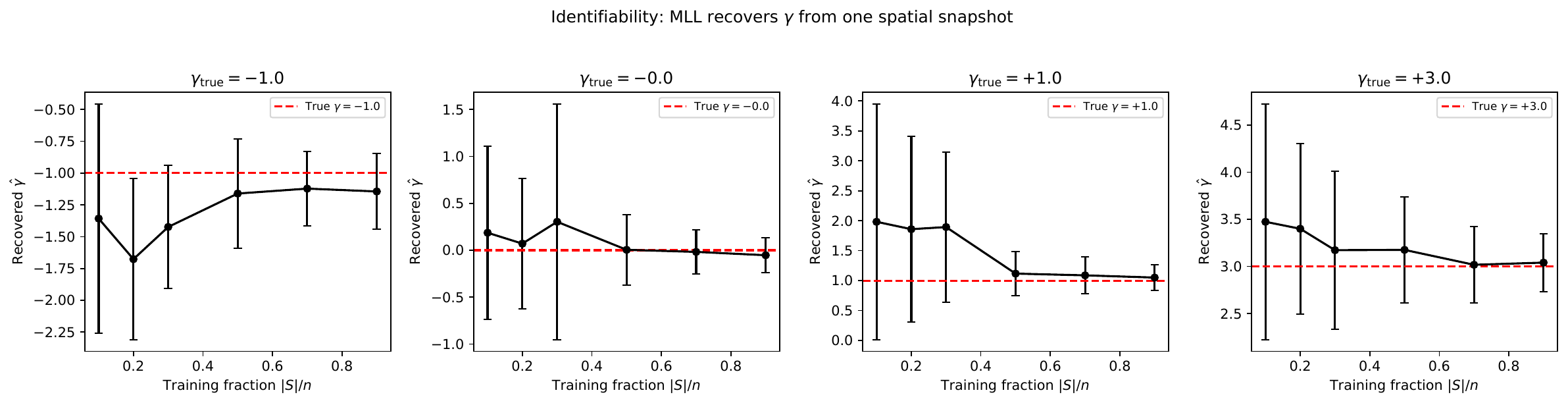}
\caption{Identifiability on a $20 \times 20$ regular grid. Recovered
$\hat\gamma$ (mean $\pm$ standard deviation over $10$ seeds) versus
training fraction $|\m{S}|/n$, for each of four ground truths
$\gamma_{\mathrm{true}} \in \{-1, 0, +1, +3\}$. The dashed line marks
$\hat\gamma = \gamma_{\mathrm{true}}$. Recovery is within $0.2$ at
$|\m{S}|/n \geq 0.5$ for every ground truth.}
\label{fig:synth-iden}
\end{figure}

\paragraph{Result.}
Recovery is within $0.2$ of the truth at $|\m{S}|/n \geq 0.5$ for every
$\gamma_{\mathrm{true}}$ tested (Figure~\ref{fig:synth-iden}). At
smaller training fractions $\hat\gamma$ overshoots in magnitude --- it lands
\emph{further} from the stationary point than the truth, not nearer it, on every
$\gamma_{\mathrm{true}}$ tested (at $\gamma_{\mathrm{true}} = -1$ the mean at
$|\m{S}|/n = 0.2$ is $-1.68$) --- and the bias decays as $|\m{S}|/n$ grows.
The joint MLL is identifiable in $\gamma$ on a graph with non-constant
$d_i$.

\subsection{Two-regime stochastic block model}
\label{sec:appendix-synth-tworegime}

\paragraph{Setup.}
A controlled two-regime stochastic block model: $n = 200$, $100$
vertices per block; sparse block A ($p_{\mathrm{in}} = 0.04$, expected
degree $\approx 4$), dense block B ($p_{\mathrm{in}} = 0.20$, expected
degree $\approx 20$), $p_{\mathrm{between}} = 0.005$; noise
$\sigma^2 = 0.05$; ground truths $\gamma_\m{A} = -1$ (sparse-block
propagation) and $\gamma_\m{B} = +2$ (dense-block isolation).

\begin{figure}[h]
\centering
\includegraphics[width=0.85\linewidth]{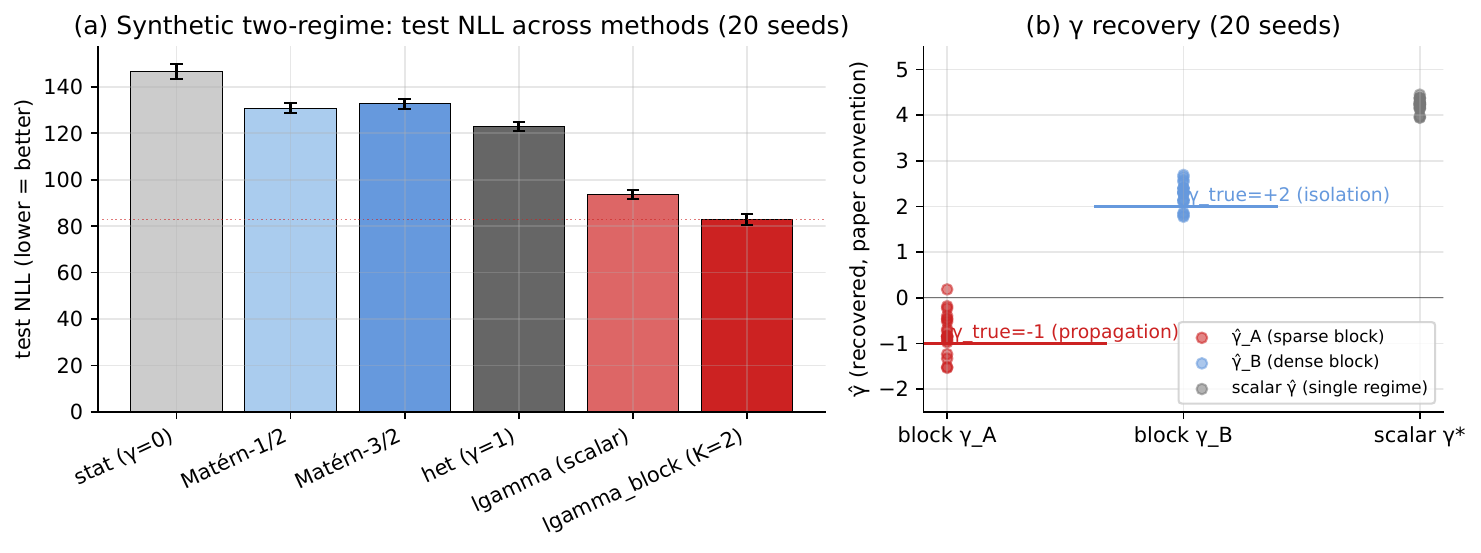}
\caption{Two-regime SBM. \emph{Left:} per-method test NLL/pt.
\emph{Right:} recovered $\hat\gamma_k$ vs.\ ground truth.}
\label{fig:synth-tworegime}
\end{figure}

\paragraph{Result.}
Six methods are run. The stationary Mat\'ern baselines \texttt{m05} and \texttt{m15} pay
an irreducible $+0.80$ and $+0.83$ nat/pt against the leader because no scalar exponent
fits both blocks, and the stationary family members bracket them
(\texttt{het} $+0.67$, \texttt{stat} $+1.06$). Scalar \texttt{lgamma} collapses to a
single compromise $\hat\gamma \approx +4.2$ and places second of the six at
$1.561$. \texttt{lgamma\_block}
($K = 2$, true block labels) recovers $\hat\gamma_\m{A} = -0.77 \pm 0.10$,
$\hat\gamma_\m{B} = +2.24 \pm 0.06$ with the correct signs and within
$0.25$ of each ground truth, and takes top-1 NLL/pt at $1.380$,
$0.18$ nat/pt clear of the next best
(Figure~\ref{fig:synth-tworegime}). The piecewise extension is
therefore necessary, not ornamental: scalar models cannot fit a
multi-regime truth.

% ==================================================================
\section{Full per-dataset leaderboards with SE and significance}
\label{sec:appendix-fullboards}
% ==================================================================

The main paper reports the top five rows of each leaderboard. The
tables below give the full $20$-seed leaderboard with mean
$\pm$ standard error (SE = std$/\sqrt{20}$) on all three metrics, plus
a paired one-sided $t$-test $p$-value asking whether the dataset
leader has significantly lower NLL/pt than the row. SE captures
seed-to-seed variability; $p < 0.05$ marks a row that the leader
beats with high confidence on raw NLL/pt. Rows with $p > 0.05$ are
within the leader's noise band on NLL/pt; some of those still lose
on RMSE or recNLL/pt (or vice versa). On METR-LA the NLL leader is
\texttt{lgamma\_fisher}, with scalar \texttt{lgamma} inside its noise band
($p = 0.44$) and \texttt{lgamma\_fisher\_m15} taking RMSE --- a Pareto split. On
PEMS-BAY the NLL leader is \texttt{lgamma\_fisher}; \texttt{het} is within its noise
band ($p = 0.17$) and has the lowest RMSE.

\begin{table}[h]
\centering
\caption{METR-LA: full $20$-seed leaderboard with mean $\pm$ SE; $p$-values from a paired one-sided $t$-test asking whether the dataset leader has significantly lower NLL/pt than the row. The $p$ column is the test against the dataset leader; the last column flags whether the row is a family member or an outer baseline.}
\label{tab:metrla_se}
\footnotesize
\setlength{\tabcolsep}{2.5pt}
\begin{tabular}{@{}l c c c c l@{}}
\toprule
Method & NLL/pt & RMSE & recNLL/pt & $p$ vs.\ leader & cat \\
\midrule
\texttt{lgamma\_fisher} & $1.2018 \pm 0.0287$ & $0.8369 \pm 0.0238$ & $1.1763 \pm 0.0232$ & --- & family (leader) \\
\texttt{lgamma} & $1.2025 \pm 0.0294$ & $0.8468 \pm 0.0247$ & $1.1759 \pm 0.0228$ & $0.437$ & family \\
\texttt{het} & $1.2201 \pm 0.0265$ & $0.8469 \pm 0.0231$ & $1.1975 \pm 0.0234$ & $0.002$ & family \\
\texttt{lgamma\_fisher\_m15} & $1.2202 \pm 0.0344$ & $0.8265 \pm 0.0235$ & $1.1816 \pm 0.0247$ & $0.049$ & family \\
\texttt{lgamma\_m15} & $1.2234 \pm 0.0349$ & $0.8354 \pm 0.0240$ & $1.1835 \pm 0.0244$ & $0.039$ & family \\
\texttt{fuglreg\_lam1.0} & $1.2382 \pm 0.0294$ & $0.8603 \pm 0.0232$ & $1.2064 \pm 0.0224$ & $<\!10^{-3}$ & base \\
\texttt{spectral\_bic\_lgamma\_block} & $1.2486 \pm 0.0341$ & $0.8828 \pm 0.0242$ & $1.2184 \pm 0.0254$ & $0.027$ & family \\
\texttt{fem} & $1.2537 \pm 0.0261$ & $0.8480 \pm 0.0225$ & $1.2301 \pm 0.0240$ & $<\!10^{-3}$ & family \\
\texttt{stat} & $1.2612 \pm 0.0265$ & $0.8491 \pm 0.0227$ & $1.2368 \pm 0.0244$ & $<\!10^{-3}$ & family \\
\texttt{m15} & $1.2790 \pm 0.0301$ & $0.8302 \pm 0.0222$ & $1.2448 \pm 0.0243$ & $<\!10^{-3}$ & base \\
\texttt{m05} & $1.2881 \pm 0.0255$ & $0.8802 \pm 0.0237$ & $1.2669 \pm 0.0238$ & $<\!10^{-3}$ & base \\
\texttt{lnu} & $1.2939 \pm 0.0330$ & $0.8307 \pm 0.0224$ & $1.2515 \pm 0.0244$ & $<\!10^{-3}$ & base \\
\texttt{se\_kriging} & $1.3289 \pm 0.0266$ & $0.8681 \pm 0.0169$ & $1.2794 \pm 0.0193$ & $<\!10^{-3}$ & base \\
\texttt{ps} & $1.5411 \pm 0.0705$ & $0.9239 \pm 0.0213$ & $1.3306 \pm 0.0257$ & $<\!10^{-3}$ & base \\
\texttt{specpoly} & $1.7334 \pm 0.0256$ & $1.2427 \pm 0.0292$ & $1.6915 \pm 0.0330$ & $<\!10^{-3}$ & base \\
\texttt{fuglstad} & $1.8988 \pm 0.1073$ & $0.9213 \pm 0.0267$ & $1.2671 \pm 0.0264$ & $<\!10^{-3}$ & base \\
\texttt{dk} & $2.7214 \pm 0.2608$ & $1.1237 \pm 0.0296$ & $1.7718 \pm 0.0441$ & $<\!10^{-3}$ & learned baseline \\
\bottomrule
\end{tabular}
\end{table}

\begin{table}[h]
\centering
\caption{PEMS-BAY: full $20$-seed leaderboard with mean $\pm$ SE; $p$-values from a paired one-sided $t$-test asking whether the dataset leader has significantly lower NLL/pt than the row. The $p$ column is the test against the dataset leader; the last column flags whether the row is a family member or an outer baseline.}
\label{tab:pemsbay_se}
\footnotesize
\setlength{\tabcolsep}{2.5pt}
\begin{tabular}{@{}l c c c c l@{}}
\toprule
Method & NLL/pt & RMSE & recNLL/pt & $p$ vs.\ leader & cat \\
\midrule
\texttt{lgamma\_fisher} & $1.2720 \pm 0.0346$ & $0.9015 \pm 0.0360$ & $1.2282 \pm 0.0371$ & --- & family (leader) \\
\texttt{het} & $1.2767 \pm 0.0329$ & $0.8906 \pm 0.0329$ & $1.2344 \pm 0.0362$ & $0.166$ & family \\
\texttt{lgamma} & $1.2835 \pm 0.0370$ & $0.9019 \pm 0.0360$ & $1.2374 \pm 0.0381$ & $0.010$ & family \\
\texttt{lgamma\_fisher\_m15} & $1.2911 \pm 0.0359$ & $0.8963 \pm 0.0329$ & $1.2459 \pm 0.0371$ & $<\!10^{-3}$ & family \\
\texttt{spectral\_bic\_lgamma\_block} & $1.2934 \pm 0.0330$ & $0.9249 \pm 0.0359$ & $1.2582 \pm 0.0329$ & $0.009$ & family \\
\texttt{lgamma\_m15} & $1.2965 \pm 0.0366$ & $0.8965 \pm 0.0329$ & $1.2504 \pm 0.0371$ & $<\!10^{-3}$ & family \\
\texttt{fem} & $1.3245 \pm 0.0320$ & $0.8908 \pm 0.0323$ & $1.2798 \pm 0.0360$ & $<\!10^{-3}$ & family \\
\texttt{fuglreg\_lam1.0} & $1.3340 \pm 0.0365$ & $0.8948 \pm 0.0331$ & $1.2918 \pm 0.0349$ & $<\!10^{-3}$ & base \\
\texttt{m05} & $1.3556 \pm 0.0356$ & $0.8960 \pm 0.0349$ & $1.3025 \pm 0.0397$ & $<\!10^{-3}$ & base \\
\texttt{lnu} & $1.3562 \pm 0.0349$ & $0.8961 \pm 0.0341$ & $1.3048 \pm 0.0388$ & $<\!10^{-3}$ & base \\
\texttt{stat} & $1.3594 \pm 0.0331$ & $0.8949 \pm 0.0326$ & $1.3129 \pm 0.0364$ & $<\!10^{-3}$ & family \\
\texttt{m15} & $1.3655 \pm 0.0352$ & $0.9016 \pm 0.0347$ & $1.3133 \pm 0.0386$ & $<\!10^{-3}$ & base \\
\texttt{se\_kriging} & $1.3867 \pm 0.0321$ & $0.9350 \pm 0.0333$ & $1.3372 \pm 0.0360$ & $<\!10^{-3}$ & base \\
\texttt{specpoly} & $1.6527 \pm 0.0327$ & $1.0999 \pm 0.0409$ & $1.5749 \pm 0.0453$ & $<\!10^{-3}$ & base \\
\texttt{fuglstad} & $1.7990 \pm 0.1087$ & $0.9181 \pm 0.0346$ & $1.3485 \pm 0.0329$ & $<\!10^{-3}$ & base \\
\texttt{ps} & $1.8058 \pm 0.0706$ & $1.0856 \pm 0.0258$ & $1.5305 \pm 0.0267$ & $<\!10^{-3}$ & base \\
\texttt{dk} & $3.1423 \pm 0.3538$ & $1.2189 \pm 0.0431$ & $1.9880 \pm 0.0520$ & $<\!10^{-3}$ & learned baseline \\
\bottomrule
\end{tabular}
\end{table}

\begin{table}[h]
\centering
\caption{LAQN: full $20$-seed leaderboard with mean $\pm$ SE; $p$-values from a paired one-sided $t$-test asking whether the dataset leader has significantly lower NLL/pt than the row. The $p$ column is the test against the dataset leader; the last column flags whether the row is a family member or an outer baseline.}
\label{tab:laqn_se}
\footnotesize
\setlength{\tabcolsep}{2.5pt}
\begin{tabular}{@{}l c c c c l@{}}
\toprule
Method & NLL/pt & RMSE & recNLL/pt & $p$ vs.\ leader & cat \\
\midrule
\texttt{lgamma\_m15} & $1.4519 \pm 0.0374$ & $1.0455 \pm 0.0383$ & $1.4068 \pm 0.0332$ & --- & family (leader) \\
\texttt{lgamma} & $1.4688 \pm 0.0400$ & $1.0434 \pm 0.0393$ & $1.4171 \pm 0.0346$ & $0.001$ & family \\
\texttt{lgamma\_fisher\_m15} & $1.4741 \pm 0.0410$ & $1.0468 \pm 0.0386$ & $1.4227 \pm 0.0355$ & $<\!10^{-3}$ & family \\
\texttt{lgamma\_fisher} & $1.4742 \pm 0.0403$ & $1.0460 \pm 0.0392$ & $1.4223 \pm 0.0347$ & $<\!10^{-3}$ & family \\
\texttt{spectral\_bic\_lgamma\_block} & $1.4779 \pm 0.0528$ & $1.0259 \pm 0.0396$ & $1.4020 \pm 0.0372$ & $0.142$ & family \\
\texttt{het} & $1.5234 \pm 0.0419$ & $1.0717 \pm 0.0356$ & $1.4676 \pm 0.0343$ & $<\!10^{-3}$ & family \\
\texttt{m05} & $1.5244 \pm 0.0487$ & $1.0447 \pm 0.0397$ & $1.4495 \pm 0.0411$ & $<\!10^{-3}$ & base \\
\texttt{m15} & $1.5262 \pm 0.0488$ & $1.0444 \pm 0.0397$ & $1.4507 \pm 0.0412$ & $<\!10^{-3}$ & base \\
\texttt{se\_kriging} & $1.5514 \pm 0.0551$ & $1.0518 \pm 0.0394$ & $1.4586 \pm 0.0395$ & $<\!10^{-3}$ & base \\
\texttt{specpoly} & $1.5524 \pm 0.0407$ & $1.1300 \pm 0.0338$ & $1.5029 \pm 0.0305$ & $<\!10^{-3}$ & base \\
\texttt{lnu} & $1.5568 \pm 0.0486$ & $1.0505 \pm 0.0358$ & $1.4796 \pm 0.0380$ & $<\!10^{-3}$ & base \\
\texttt{fem} & $1.5590 \pm 0.0446$ & $1.0624 \pm 0.0344$ & $1.4917 \pm 0.0366$ & $<\!10^{-3}$ & family \\
\texttt{fuglreg\_lam1.0} & $1.5895 \pm 0.0555$ & $1.0573 \pm 0.0373$ & $1.4973 \pm 0.0421$ & $<\!10^{-3}$ & base \\
\texttt{stat} & $1.6013 \pm 0.0487$ & $1.0634 \pm 0.0340$ & $1.5188 \pm 0.0381$ & $<\!10^{-3}$ & family \\
\texttt{ps} & $2.2013 \pm 0.1941$ & $1.2804 \pm 0.0895$ & $1.7665 \pm 0.0521$ & $<\!10^{-3}$ & base \\
\texttt{fuglstad} & $2.3218 \pm 0.2453$ & $1.1173 \pm 0.0442$ & $1.5674 \pm 0.0495$ & $<\!10^{-3}$ & base \\
\texttt{dk} & $4.3435 \pm 0.2683$ & $1.2220 \pm 0.0448$ & $1.6085 \pm 0.0419$ & $<\!10^{-3}$ & learned baseline \\
\bottomrule
\end{tabular}
\end{table}

\begin{figure}[!h]
\centering
\includegraphics[width=0.55\linewidth]{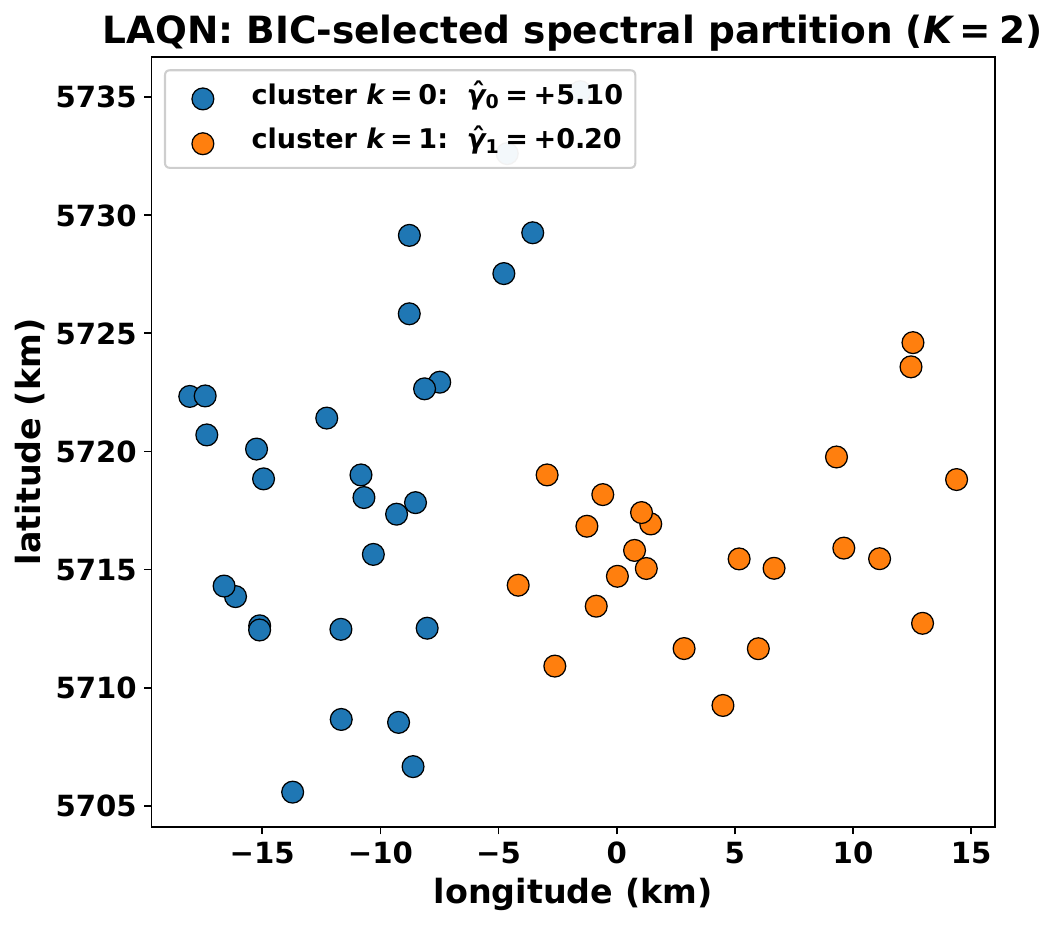}
\caption{LAQN spectral partition. Sensors are coloured by the
two-cluster spectral partition of $\m{L}$ chosen by BIC; the
per-cluster $\hat\gamma_k$ values are shown. The partition is geographic rather than by
site type: the two clusters carry almost the same mix of site classes (mean site-type
ordinal $1.13$ against $1.06$) and separate east from west.}
\label{fig:laqn_leader}
\end{figure}

\begin{table}[h]
\centering
\caption{PM2.5: full $20$-seed leaderboard with mean $\pm$ SE; $p$-values from a paired one-sided $t$-test asking whether the dataset leader has significantly lower NLL/pt than the row. The $p$ column is the test against the dataset leader; the last column flags whether the row is a family member or an outer baseline.}
\label{tab:pm25_se}
\footnotesize
\setlength{\tabcolsep}{1.5pt}
\begin{tabular}{@{}l c c c c l@{}}
\toprule
Method & NLL/pt & RMSE & recNLL/pt & $p$ & cat \\
\midrule
\texttt{lgamma\_block\_aug}\,($K{=}20, w{=}1$) & $0.9756 \pm 0.0106$ & $0.6541 \pm 0.0055$ & $0.9703 \pm 0.0095$ & --- & family (ldr) \\
\texttt{lgamma\_block\_aug}\,($K{=}10, w{=}1$) & $1.0078 \pm 0.0105$ & $0.6822 \pm 0.0075$ & $1.0032 \pm 0.0099$ & $<\!10^{-3}$ & family \\
\texttt{lgamma\_block\_aug}\,($K{=}10, w{=}0.5$) & $1.0303 \pm 0.0097$ & $0.7011 \pm 0.0069$ & $1.0269 \pm 0.0090$ & $<\!10^{-3}$ & family \\
\texttt{lgamma\_block\_aug}\,($K{=}20, w{=}0.5$) & $1.0410 \pm 0.0121$ & $0.6975 \pm 0.0072$ & $1.0338 \pm 0.0102$ & $<\!10^{-3}$ & family \\
\texttt{lgamma\_block\_aug}\,($K{=}5, w{=}1$) & $1.0597 \pm 0.0099$ & $0.7171 \pm 0.0069$ & $1.0553 \pm 0.0095$ & $<\!10^{-3}$ & family \\
\texttt{lgamma\_block\_aug}\,($K{=}5, w{=}0.5$) & $1.0698 \pm 0.0082$ & $0.7323 \pm 0.0062$ & $1.0672 \pm 0.0080$ & $<\!10^{-3}$ & family \\
\texttt{fuglreg\_lam1.0} & $1.0748 \pm 0.0095$ & $0.7467 \pm 0.0062$ & $1.0705 \pm 0.0082$ & $<\!10^{-3}$ & base \\
\texttt{lgamma\_block\_aug}\,($K{=}2, w{=}1$) & $1.0803 \pm 0.0114$ & $0.7012 \pm 0.0070$ & $1.0752 \pm 0.0107$ & $<\!10^{-3}$ & family \\
\texttt{m15} & $1.1194 \pm 0.0087$ & $0.7392 \pm 0.0062$ & $1.1164 \pm 0.0087$ & $<\!10^{-3}$ & base \\
\texttt{lnu} & $1.1229 \pm 0.0084$ & $0.7417 \pm 0.0061$ & $1.1200 \pm 0.0086$ & $<\!10^{-3}$ & base \\
\texttt{lgamma\_block\_aug}\,($K{=}2, w{=}0.5$) & $1.1249 \pm 0.0075$ & $0.7419 \pm 0.0060$ & $1.1225 \pm 0.0074$ & $<\!10^{-3}$ & family \\
\texttt{lgamma} & $1.1261 \pm 0.0086$ & $0.7419 \pm 0.0059$ & $1.1230 \pm 0.0087$ & $<\!10^{-3}$ & family \\
\texttt{spectral\_bic\_lgamma\_block} & $1.1312 \pm 0.0114$ & $0.7415 \pm 0.0073$ & $1.1237 \pm 0.0099$ & $<\!10^{-3}$ & family \\
\texttt{stat} & $1.1329 \pm 0.0080$ & $0.7484 \pm 0.0060$ & $1.1297 \pm 0.0085$ & $<\!10^{-3}$ & family \\
\texttt{m05} & $1.1358 \pm 0.0079$ & $0.7517 \pm 0.0063$ & $1.1320 \pm 0.0086$ & $<\!10^{-3}$ & base \\
\texttt{lgamma\_m15} & $1.1367 \pm 0.0086$ & $0.7489 \pm 0.0059$ & $1.1336 \pm 0.0085$ & $<\!10^{-3}$ & family \\
\texttt{lgamma\_fisher} & $1.1443 \pm 0.0089$ & $0.7520 \pm 0.0063$ & $1.1417 \pm 0.0092$ & $<\!10^{-3}$ & family \\
\texttt{lgamma\_fisher\_m15} & $1.1510 \pm 0.0088$ & $0.7588 \pm 0.0061$ & $1.1482 \pm 0.0090$ & $<\!10^{-3}$ & family \\
\texttt{fem} & $1.1557 \pm 0.0080$ & $0.7601 \pm 0.0058$ & $1.1530 \pm 0.0083$ & $<\!10^{-3}$ & family \\
\texttt{se\_kriging} & $1.1921 \pm 0.0093$ & $0.7780 \pm 0.0060$ & $1.1668 \pm 0.0077$ & $<\!10^{-3}$ & base \\
\texttt{het} & $1.2403 \pm 0.0092$ & $0.7761 \pm 0.0071$ & $1.2369 \pm 0.0096$ & $<\!10^{-3}$ & family \\
\texttt{ps} & $1.5547 \pm 0.0192$ & $0.9300 \pm 0.0081$ & $1.4806 \pm 0.0119$ & $<\!10^{-3}$ & base \\
\texttt{specpoly} & $1.6845 \pm 0.0072$ & $1.3232 \pm 0.0093$ & $1.6821 \pm 0.0068$ & $<\!10^{-3}$ & base \\
\texttt{dk} & $2.1634 \pm 0.0805$ & $0.9692 \pm 0.0131$ & $1.6526 \pm 0.0211$ & $<\!10^{-3}$ & learned baseline \\
\texttt{fuglstad} & $2.2946 \pm 0.0641$ & $0.7655 \pm 0.0064$ & $1.1754 \pm 0.0103$ & $<\!10^{-3}$ & base \\
\bottomrule
\end{tabular}
\end{table}

\begin{figure}[!h]
\centering
\includegraphics[width=0.85\linewidth]{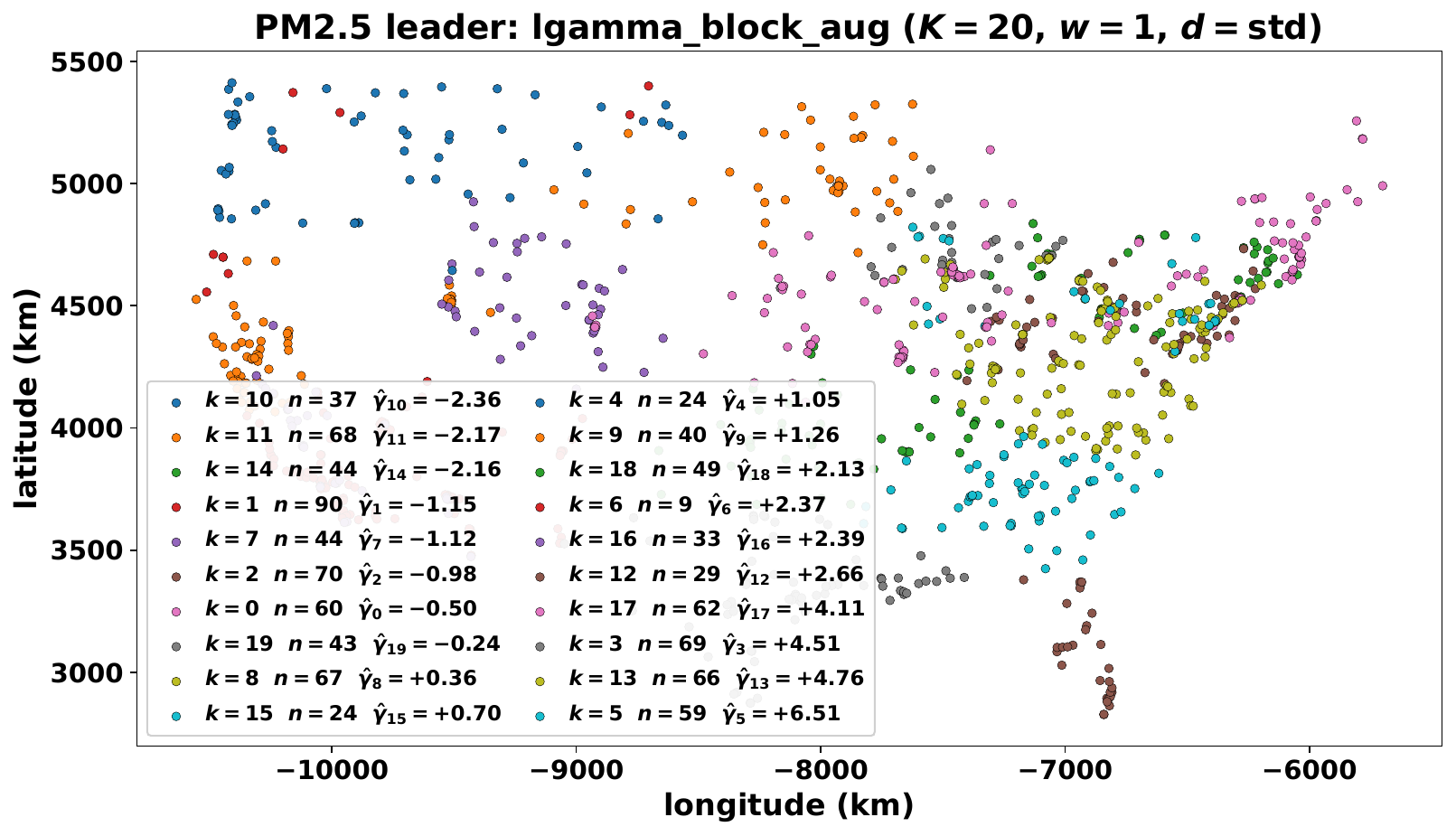}
\caption{PM2.5 spatial leader. Stations are coloured by the augmented
$K$-means partition ($K = 20$, $w = 1$); the per-cluster $\hat\gamma_k$ values are shown.
The fitted exponents span both regimes---propagation on some clusters, isolation on
others---which is what a single global $\gamma$ cannot represent.}
\label{fig:pm25_leader}
\end{figure}

\begin{table}[h]
\centering
\caption{CA Weather: $20$-seed leaderboard including the Fisher, Mat\'ern and spectral-BIC members; only the aggregate-best augmented configuration is shown (its $K$-grid is in Table~\ref{tab:kgrid-supp}). Each entry is the mean over the $12$ monthly snapshots, with mean $\pm$ SE across seeds; $p$-values from a paired one-sided $t$-test asking whether the dataset leader has significantly lower NLL/pt than the row. The $p$ column is the test against the dataset leader; the last column flags whether the row is a family member or an outer baseline.}
\label{tab:caweather_se}
\footnotesize
\setlength{\tabcolsep}{2.5pt}
\begin{tabular}{@{}l c c c c l@{}}
\toprule
Method & NLL/pt & RMSE & recNLL/pt & $p$ & cat \\
\midrule
\texttt{lgamma\_block\_aug}\textsubscript{$K{=}10,w{=}1$} & $0.8685 \pm 0.0079$ & $0.5738 \pm 0.0049$ & $0.8572 \pm 0.0076$ & --- & family (ldr) \\
\texttt{ps} & $0.9360 \pm 0.0088$ & $0.6261 \pm 0.0053$ & $0.9247 \pm 0.0078$ & $<\!10^{-3}$ & base \\
\texttt{se\_kriging} & $0.9511 \pm 0.0061$ & $0.6245 \pm 0.0043$ & $0.9376 \pm 0.0065$ & $<\!10^{-3}$ & base \\
\texttt{spectral\_bic\_lgamma\_block} & $0.9562 \pm 0.0229$ & $0.6151 \pm 0.0094$ & $0.9364 \pm 0.0163$ & $<\!10^{-3}$ & family \\
\texttt{fuglreg\_lam1.0} & $1.0004 \pm 0.0087$ & $0.6451 \pm 0.0055$ & $0.9900 \pm 0.0076$ & $<\!10^{-3}$ & base \\
\texttt{lgamma} & $1.0031 \pm 0.0086$ & $0.6562 \pm 0.0069$ & $0.9910 \pm 0.0089$ & $<\!10^{-3}$ & family \\
\texttt{lgamma\_m15} & $1.0184 \pm 0.0093$ & $0.6706 \pm 0.0070$ & $1.0074 \pm 0.0092$ & $<\!10^{-3}$ & family \\
\texttt{lgamma\_fisher} & $1.0190 \pm 0.0087$ & $0.6556 \pm 0.0063$ & $1.0074 \pm 0.0090$ & $<\!10^{-3}$ & family \\
\texttt{lnu} & $1.0346 \pm 0.0077$ & $0.6522 \pm 0.0063$ & $1.0224 \pm 0.0083$ & $<\!10^{-3}$ & base \\
\texttt{m05} & $1.0352 \pm 0.0072$ & $0.6596 \pm 0.0059$ & $1.0220 \pm 0.0079$ & $<\!10^{-3}$ & base \\
\texttt{m15} & $1.0375 \pm 0.0076$ & $0.6597 \pm 0.0056$ & $1.0258 \pm 0.0079$ & $<\!10^{-3}$ & base \\
\texttt{stat} & $1.0436 \pm 0.0061$ & $0.6551 \pm 0.0049$ & $1.0317 \pm 0.0067$ & $<\!10^{-3}$ & family \\
\texttt{lgamma\_fisher\_m15} & $1.0521 \pm 0.0092$ & $0.6801 \pm 0.0064$ & $1.0420 \pm 0.0091$ & $<\!10^{-3}$ & family \\
\texttt{fem} & $1.0673 \pm 0.0061$ & $0.6576 \pm 0.0050$ & $1.0568 \pm 0.0066$ & $<\!10^{-3}$ & family \\
\texttt{het} & $1.1661 \pm 0.0055$ & $0.6917 \pm 0.0047$ & $1.1556 \pm 0.0060$ & $<\!10^{-3}$ & family \\
\texttt{specpoly} & $1.7562 \pm 0.0073$ & $1.5168 \pm 0.0099$ & $1.7448 \pm 0.0084$ & $<\!10^{-3}$ & base \\
\texttt{dk} & $1.9763 \pm 0.0689$ & $0.9714 \pm 0.0089$ & $1.6279 \pm 0.0281$ & $<\!10^{-3}$ & learned baseline \\
\texttt{fuglstad} & $3.2935 \pm 0.0758$ & $0.8032 \pm 0.0120$ & $1.2211 \pm 0.0089$ & $<\!10^{-3}$ & base \\
\bottomrule
\end{tabular}
\end{table}

% ==================================================================
\section{Deep Ensemble GNN comparison}
\label{sec:appendix-degnn}
% ==================================================================

\paragraph{Question.}
Deep ensembles are a standard, general-purpose baseline for predictive
uncertainty~\citep{Lakshminarayanan2017}, so we include one among our baselines
here---a deep ensemble of Graph Convolutional Networks~\citep{KipfWelling2017}
with mean and variance heads---even though whether such ensembles faithfully
capture uncertainty on graph-structured data has recently been called into
question~\citep{Vieira2026}.
Does a neural baseline
of this kind close the gap on the family, particularly on calibration
(measured by the recalibration scalar $c^\star$ that minimises NLL)?

\paragraph{What this table compares.}
The rows below put the deep ensemble against scalar \texttt{lgamma}, the simplest member
of the family and not the reported leader on any dataset, which makes it the hardest
comparison for us: on CA Weather the ensemble is ahead of scalar \texttt{lgamma} on seven
of the twelve months, significantly so on four and significantly behind on two, and over
the twelve-month mean the two are indistinguishable ($0.9963$ against $1.0031$,
$p = 0.55$). That is the expected place for it to be competitive, CA Weather being the
dataset a single global exponent cannot fit. Against the \emph{dataset leader} the
ensemble loses \emph{on raw NLL} on all five datasets, by $0.13$ nat/pt on CA Weather and
$0.17$ on PM2.5 up to $3.97$ on LAQN, with worst-case $p = 0.008$, and on RMSE as well
(METR-LA $0.9232$ against $0.8265$, $p = 0.003$); the deep kernel loses by $1.11$ to
$2.89$. Recalibrated NLL is the one metric on which this is not uniformly true, and
METR-LA is the one dataset: see below. Those comparisons are computed by
\texttt{graph\_gp/compare\_learned\_baselines.py}.

\paragraph{Setup.}
We train $M = 5$ independent GCNs ($3$ graph-convolution layers, $64$ hidden units,
$300$ epochs, Adam $\mathrm{lr} = 10^{-2}$) on the training mask, each with a
mean and a log-variance output head and a Gaussian NLL loss. Each node carries two input
features, the auxiliary field $d_i$ and the log-degree, so the network sees the same
input information as the \texttt{lgamma} family. The
ensemble prediction averages means and combines within-model and
between-model variance in the standard way. We report raw NLL/pt,
recalibrated NLL/pt (which fits a single test-side scalar $c^\star$),
and $c^\star$ itself; $c^\star \approx 1$ is well-calibrated, $c^\star
> 1$ is overconfident, $c^\star < 1$ is underconfident. The paired
one-sided $t$-test asks whether the Deep Ensemble has \emph{larger}
NLL than \texttt{lgamma}; a dagger marks the reverse direction.

\begin{table}[h]
\centering
\caption{Deep Ensemble GNN ($M = 5$) versus \texttt{lgamma} across all
$16$ settings, $20$ seeds each. NLL is per-test-point. $c^{\star} > 1$
means the predictive variance is too small (overconfident).}
\label{tab:degnn}
\footnotesize
\setlength{\tabcolsep}{2pt}
\begin{tabular}{@{}lcccccccr@{}}
\toprule
Setting & $\mathrm{NLL}_{\mathrm{DE}}$ & $c^{\star}_{\mathrm{DE}}$
        & $\mathrm{NLL}_{\mathrm{lg}}$ & $c^{\star}_{\mathrm{lg}}$
        & $\mathrm{rec}_{\mathrm{DE}}$ & $\mathrm{rec}_{\mathrm{lg}}$
        & $\Delta$ & $p$(DE $>$ lg) \\
\midrule
METR-LA       & $1.5820 \pm 0.1465$ & $2.73$ & $1.2025 \pm 0.0294$ & $1.14$ & $1.1417$ & $1.1759$ & $+0.3796$ & $\phantom{<}0.009^{*}$ \\
PEMS-BAY      & $1.6402 \pm 0.0914$ & $2.49$ & $1.2835 \pm 0.0370$ & $1.09$ & $1.3196$ & $1.2374$ & $+0.3567$ & $<0.001^{*}$ \\
LAQN          & $5.4225 \pm 0.9784$ & $9.97$ & $1.4688 \pm 0.0400$ & $1.20$ & $1.9130$ & $1.4171$ & $+3.9537$ & $<0.001^{*}$ \\
PM2.5         & $1.1474 \pm 0.0129$ & $1.27$ & $1.1261 \pm 0.0086$ & $1.00$ & $1.1276$ & $1.1230$ & $+0.0213$ & $\phantom{<}0.020^{*}$ \\
CAW Jan       & $0.9832 \pm 0.0241$ & $1.29$ & $0.9718 \pm 0.0169$ & $0.92$ & $0.9574$ & $0.9588$ & $+0.0114$ & $\phantom{<}0.270$ \\
CAW Feb       & $0.8080 \pm 0.0164$ & $1.25$ & $0.7692 \pm 0.0129$ & $0.91$ & $0.7894$ & $0.7620$ & $+0.0388$ & $\phantom{<}0.011^{*}$ \\
CAW Mar       & $0.9166 \pm 0.0136$ & $1.32$ & $0.9206 \pm 0.0111$ & $0.93$ & $0.8921$ & $0.9144$ & $-0.0040$ & $\phantom{<}0.615$ \\
CAW Apr       & $1.0932 \pm 0.0209$ & $1.30$ & $1.0850 \pm 0.0121$ & $0.93$ & $1.0646$ & $1.0776$ & $+0.0082$ & $\phantom{<}0.298$ \\
CAW May       & $1.0433 \pm 0.0189$ & $1.32$ & $1.0755 \pm 0.0087$ & $0.91$ & $1.0162$ & $1.0698$ & $-0.0322$ & $\phantom{<}0.975^{\dagger}$ \\
CAW Jun       & $0.9391 \pm 0.0192$ & $1.27$ & $0.9848 \pm 0.0078$ & $0.88$ & $0.9180$ & $0.9765$ & $-0.0457$ & $\phantom{<}0.991^{\dagger}$ \\
CAW Jul       & $1.1484 \pm 0.0345$ & $1.47$ & $1.1030 \pm 0.0146$ & $0.94$ & $1.0949$ & $1.0940$ & $+0.0454$ & $\phantom{<}0.048^{*}$ \\
CAW Aug       & $1.1287 \pm 0.0295$ & $1.38$ & $1.1085 \pm 0.0113$ & $0.91$ & $1.0883$ & $1.1007$ & $+0.0202$ & $\phantom{<}0.190$ \\
CAW Sep       & $1.0165 \pm 0.0217$ & $1.28$ & $1.0521 \pm 0.0139$ & $0.89$ & $0.9926$ & $1.0398$ & $-0.0356$ & $\phantom{<}0.976^{\dagger}$ \\
CAW Oct       & $1.0471 \pm 0.0128$ & $1.23$ & $1.0837 \pm 0.0071$ & $1.00$ & $1.0320$ & $1.0824$ & $-0.0366$ & $\phantom{<}0.998^{\dagger}$ \\
CAW Nov       & $1.1004 \pm 0.0353$ & $1.54$ & $1.1231 \pm 0.0433$ & $0.92$ & $1.0343$ & $1.0628$ & $-0.0227$ & $\phantom{<}0.879$ \\
CAW Dec       & $0.7311 \pm 0.0188$ & $1.23$ & $0.7603 \pm 0.0118$ & $0.92$ & $0.7134$ & $0.7536$ & $-0.0293$ & $\phantom{<}0.925$ \\
\bottomrule
\end{tabular}\\[0.5em]
\begin{minipage}{0.95\textwidth}\footnotesize
$^{*}$ $p < 0.05$: Deep Ensemble is significantly \emph{worse} than
\texttt{lgamma}. $^{\dagger}$ $1 - p < 0.05$: Deep Ensemble is
significantly \emph{better}. Paired one-sided $t$-test over $20$ seeds.
\end{minipage}
\end{table}

\paragraph{Inference --- sensor networks.}
On all four single-snapshot sensor datasets the Deep Ensemble is
severely overconfident: $c^\star \in \{1.27, 2.49, 2.73, 9.97\}$ on
PM2.5 / PEMS-BAY / METR-LA / LAQN. The predictive variance must be
inflated by a factor of roughly $2.5$--$10$ to match the empirical residuals on
the three smaller graphs. This is the classical overconfidence
failure of deep models trained by maximum likelihood with a variance
head: the $M$ networks converge to similar solutions on a single
snapshot, collapsing to tight variance estimates. The raw-NLL gap of
$0.02$--$3.95$ nat/pt is mostly a calibration failure. After
recalibration \texttt{lgamma} is still lower on three of the four
(PEMS-BAY $1.2374$ against $1.3196$, LAQN $1.4171$ against $1.9130$, PM2.5 $1.1230$
against $1.1276$), but METR-LA is a genuine exception and we state it plainly: there the
recalibrated ensemble reaches $1.1417$ nat/pt, nominally the lowest of any method we run,
below the family's best recalibrated value of $1.1759$. The difference is not significant
against any family member --- paired $p = 0.47$ against \texttt{lgamma}, $0.44$ against
\texttt{lgamma\_fisher}, $0.21$ against \texttt{het}, and significant only against the
stationary baseline \texttt{stat} ($p = 0.029$). The reading is that on METR-LA the
ensemble's raw-NLL deficit is almost entirely a single global variance scale
($c^\star = 2.73$) rather than a worse mean or a worse variance \emph{shape}, and that
the scalar which would close it is fitted on the test set and so is not available at
deployment time. This is the one place in the paper where a learned baseline is not
behind once miscalibration is removed.

\paragraph{The other learned-representation baseline: a deep-kernel GP.}
We also run a deep-kernel Gaussian process---a learned feature map with a stationary
kernel in feature space---reported as \texttt{dk} in the per-dataset leaderboards. It is
the weaker of the two learned representations by a wide margin: NLL/pt $2.1634$ and RMSE
$0.9692$ on PM2.5 against the family's $0.9756$ and $0.6541$, and RMSE $1.1237$ on
METR-LA against the family's $0.8265$. The pattern on the sensor graphs is the same one
the Deep Ensemble shows: both are flexible, high-capacity learned representations, and
where the observations number far fewer than the nodes they overfit relative to a
low-parameter fixed-form kernel. This is the comparison between a learned representation
and a structured prior that the one-scalar construction is designed to win in the
small-data regime, and it is why we report a fixed-form kernel rather than a learned
feature map.

\paragraph{Inference --- CA Weather.}
On the twelve CA Weather months the picture changes. The Deep Ensemble
wins significantly on $4$ of $12$ months, loses significantly on $2$,
and ties on the remaining $6$. Its $c^\star$ values cluster near
$1.30$ --- still overconfident, but within an order of magnitude ---
while \texttt{lgamma}'s $c^\star$ sits near $0.9$ (very slightly
underconfident). CA Weather has enough spatial structure in the mean
(orography, marine-layer gradients) for a 2-layer GCN to extract
signal that a shallow GP does not; on sensor networks, where nodes
are a sparse sample of a traffic field, the GP with non-stationary
coupling wins decisively. We keep the Deep Ensemble GNN as a
calibration baseline that complements the Mat\'ern family rather
than folding it into the main leaderboards.

% ==================================================================
\section{Hyperparameters and numerical settings}
\label{sec:appendix-hyper}
% ==================================================================

All experiments use Adam with default $\beta_1 = 0.9$, $\beta_2 = 0.999$.
Hyperparameters are learned in log-space (positivity by reparameterisation).
Adam learning rate $10^{-2}$ for sensor networks (METR-LA, PEMS-BAY,
LAQN), $5\times 10^{-3}$ for PM2.5 and CA Weather. Iteration count
$T_{\mathrm{iter}} \in \{200, 300\}$; convergence is monitored by
relative change of MLL ($< 10^{-6}$ over the last $20$ steps).

\paragraph{Variance field clipping.}
$d_i$ is normalised to mean $1$ and clipped to $[0.1, 10]$ for two
numerical reasons: (a) the diagonal of $Q_\gamma$ contains the term $D^{-1}$,
so an unbounded $d_i \to 0$ would create an arbitrarily large diagonal
entry and worsen $\mathrm{cond}(Q_\gamma)$ by orders of magnitude;
(b) the off-diagonal coupling carries $d_i^{-\gamma/2} d_j^{-\gamma/2}$,
so an unbounded $d_i \to \infty$ would zero out coupling on that node
when $\hat\gamma$ is large positive (e.g.\ $\hat\gamma \approx +3$ on
METR-LA). The bracket is one decade either side of the normalised mean
--- wide enough to encode genuine heteroscedasticity, narrow enough to
keep the Cholesky stable. As a sensitivity check we re-ran the leakage-free PM2.5 leader
(\texttt{lgamma\_block\_aug} with $K = 20$, $w = 1$, $d = {}$std) on the same $20$ seeds
under two alternative brackets. Widening to $[0.05, 20]$ reproduces the default exactly
(NLL/pt $0.9756 \pm 0.0106$, RMSE $0.6541 \pm 0.0055$, with no per-seed difference at all):
the default bracket is not binding on PM2.5, since the normalised $d_i$ distribution lives
entirely within $[0.1, 10]$. Tightening to $[0.5, 2]$ shifts NLL/pt to
$0.9857 \pm 0.0107$ and RMSE to $0.6631 \pm 0.0070$, a paired deterioration of
$+0.010$ nat/pt ($p = 0.019$). The effect is therefore statistically detectable but small:
the $d_i$ values outside $[0.5, 2]$ do carry some heteroscedastic information that the
kernel exploits. This is a one-sided sensitivity result: the choice of
$[0.1, 10]$ is wide enough not to bind on the auxiliary, and tightening
it would lose information.

\paragraph{$\gamma$ bounds.}
$\gamma \in [-8, 8]$ is enforced softly via a tanh reparameterisation,
giving a smooth gradient near the bounds. They are essentially never reached: on the
PM2.5 leader $3$ of $400$ cluster exponents exceed $7.5$ in magnitude (largest $7.80$) and
on the CA Weather leader $3$ of $2400$ (largest $7.71$), so a handful come within $0.2$ of
the bound while the rest sit well inside it.

\paragraph{Float64 throughout.} All Cholesky factorisations and matrix
inversions are performed in double precision.

% ==================================================================
\section{CA Weather monthly leaderboards}
\label{sec:appendix-monthly}
% ==================================================================

The aggregate-selected augmented-$K$-means member at $(K,w)=(10,1)$
beats the outer baselines on mean NLL/pt and RMSE in each monthly snapshot.
Table~\ref{tab:caweather-monthly} compares this fixed member with the
lowest-NLL non-augmented comparator each month, including the spectral-BIC
and Fisher members. Other augmented configurations are excluded from this
comparison; the fixed member is not the monthly optimum over that grid. The thirteen outer baselines counted here are \texttt{m05}, \texttt{m15},
\texttt{lnu}, \texttt{ps}, \texttt{se\_kriging}, \texttt{dk}, \texttt{specpoly},
\texttt{fuglstad} and the five \texttt{fuglreg\_lam}$\lambda$ settings
$\lambda \in \{0.01, 0.1, 1, 10, 100\}$, each counted separately rather than collapsed
to its best value as in the leaderboards. At $p < 0.05$ the leader's significance against
\emph{all} thirteen outer
baselines holds in $8$ of $12$ months on both NLL/pt and RMSE ($9$ on NLL/pt alone,
$10$ on RMSE alone); under Bonferroni across the twelve months
($p < 0.0042 = 0.05/12$) five months survive on both, May through September. The
shortfalls are concentrated in winter, and in every case the baseline that is not
significantly beaten is \texttt{ps}, \texttt{m05}, \texttt{lnu} or
\texttt{fuglreg\_lam1.0}.

\begin{table}[h]
\centering
\caption{CA Weather: the aggregate-selected augmented member $(K,w)=(10,1)$ versus the lowest-NLL comparator each month ($20$-seed means). Comparators are the non-augmented members in Table~\ref{tab:caweather_se}, including spectral-BIC and Fisher members. Other augmented configurations are excluded; the fixed member need not be the monthly optimum over that grid. RMSE uses the same comparator selected by NLL. Negative gap favours the fixed augmented member.}
\label{tab:caweather-monthly}
\footnotesize
\setlength{\tabcolsep}{3pt}
\begin{tabular}{@{}llrrrrr@{}}
\toprule
Month & Comparator & NLL/pt aug. & NLL/pt comp. & $\Delta$NLL/pt & RMSE aug. & RMSE comp. \\
\midrule
Jan & \texttt{ps} & $0.8561$ & $0.8764$ & $-0.020$ & $0.5731$ & $0.5800$ \\
Feb & \texttt{ps} & $0.6390$ & $0.6682$ & $-0.029$ & $0.4657$ & $0.4732$ \\
Mar & \texttt{ps} & $0.8470$ & $0.8683$ & $-0.021$ & $0.5661$ & $0.5867$ \\
Apr & \texttt{se\_kriging} & $0.9889$ & $1.0292$ & $-0.040$ & $0.6353$ & $0.6773$ \\
May & \texttt{se\_kriging} & $0.8846$ & $0.9726$ & $-0.088$ & $0.5802$ & $0.6368$ \\
Jun & \texttt{spectral\_bic\_lgamma\_block} & $0.8656$ & $0.9308$ & $-0.065$ & $0.5726$ & $0.5958$ \\
Jul & \texttt{spectral\_bic\_lgamma\_block} & $0.9129$ & $1.0171$ & $-0.104$ & $0.5976$ & $0.6872$ \\
Aug & \texttt{se\_kriging} & $0.9921$ & $1.0497$ & $-0.058$ & $0.6430$ & $0.6902$ \\
Sep & \texttt{se\_kriging} & $0.8879$ & $0.9648$ & $-0.077$ & $0.5771$ & $0.6316$ \\
Oct & \texttt{ps} & $0.9106$ & $0.9293$ & $-0.019$ & $0.5949$ & $0.6102$ \\
Nov & \texttt{spectral\_bic\_lgamma\_block} & $1.0531$ & $1.0621$ & $-0.009$ & $0.6569$ & $0.6750$ \\
Dec & \texttt{ps} & $0.5841$ & $0.5922$ & $-0.008$ & $0.4225$ & $0.4335$ \\
\midrule
mean & (varies) & $0.8685$ & $0.9134$ & $-0.045$ & $0.5738$ & $0.6065$ \\
\bottomrule
\end{tabular}
\end{table}

\paragraph{Seasonal pattern.}
The lowest-NLL non-augmented comparator varies: in winter months
(Dec--Mar, Oct) the comparator is \texttt{ps} --- the
spatially-varying lengthscale GP, well-suited to the smooth
temperature fields that dominate these months, where the per-cluster
$\hat\gamma_k$ spread is narrow. Mild propagation ($\hat\gamma_k < 0$) prevails in
January, October and November ($7$, $6$ and $6$ of ten clusters), while February, April and
December split evenly and March tips the other way ($4$ of ten), so the winter pattern is a
narrow spread rather than a uniform sign. In late spring
and summer (Apr--Sep) the comparator is \texttt{se\_kriging} or
\texttt{spectral\_bic\_lgamma\_block} --- methods that exploit more
variance-field structure when the per-cluster $\hat\gamma_k$ spread
widens; July in particular carries a strong-isolation spike
($\hat\gamma_k \approx +5.9$) coexisting with the usual propagation
ranks, consistent with the seasonal $\hat\gamma_k$ heatmap shown in
the main paper (\S5, CA Weather).

% ==================================================================
\section{Scalability of the kernel factorisation}
\label{sec:appendix-scalability}
% ==================================================================

The precision $Q_\gamma = D^{-1} + 2\tau\,D^{-\gamma/2} L D^{-\gamma/2}$
inherits the sparsity pattern of $L$, so the dominant cost of the GP
fit ---  the factorisation of $Q_\gamma$ ---  reduces from
$\mathcal{O}(n^3)$ (dense inverse) to roughly $\mathcal{O}(n)$
on planar-like graphs (sparse LU on the $k$-NN or distance-threshold
adjacency). We measure this directly: for each real graph at the
canonical operating point ($\gamma = 1$, $\tau = 1$), we time the
dense inverse and the sparse LU factorisation
(\texttt{scipy.sparse.linalg.splu}) on a single CPU and report the
minimum of five runs. Table~\ref{tab:timing-supp} shows both columns.
The sparse column is the one the argument rests on, and it behaves as the construction
predicts: it grows smoothly with $n$, from $0.1$\,ms at $n = 52$ to $4.5$\,ms at
$n = 988$. The sub-millisecond LAQN entries are noise-bound. The per-dataset ratio does not
increase monotonically with $n$, for the reason given after the table.

\begin{table}[h]
\centering
\caption{Factorisation of $Q_\gamma$ for the five real graphs at
$\gamma = 1$, $\tau = 1$. Minimum of five runs on a single CPU;
``dense'' is dense inverse, ``sparse'' is sparse LU.}
\label{tab:timing-supp}
\small
\begin{tabular}{@{}lrrrrr@{}}
\toprule
Dataset & $n$ & mean deg. & dense (ms) & sparse (ms) & speed-up \\
\midrule
LAQN       &   52 &  4.9 &   0.1 &  0.1 &  1.2$\times$ \\
METR-LA    &  207 & 12.7 &  71.2 &  1.0 & 72.9$\times$ \\
PEMS-BAY   &  325 & 12.8 &   7.0 &  1.6 &  4.5$\times$ \\
CA Weather &  649 & 12.2 &  26.5 &  2.7 &  9.7$\times$ \\
PM2.5      &  988 & 10.1 &  47.2 &  4.5 & 10.4$\times$ \\
\bottomrule
\end{tabular}\\[0.3em]
\footnotesize The PM2.5 factorisation benchmark uses a $988$-vertex graph,
including one vertex excluded by the regression loader, which uses $987$
vertices. This single-vertex difference does not change the factorisation
scaling discussed here.
\end{table}

\paragraph{End-to-end cost of a fit and a prediction.}
The table above times the factorisation of $\m{Q}_\gamma$ alone. The cost a practitioner
actually pays also includes the autodiff-traced build of $\m{Q}_\gamma$, its inverse, the
backward pass, and the posterior Cholesky. At the largest operating point (PM2.5,
$n = 987$, $|\m{S}| = 690$) on a single CPU core, one Adam step costs $804$\,ms, so a full
$300$-step fit takes about $241$\,s, and prediction is a single posterior solve at
$15$\,ms. Two honest points. First, that solve is the standard exact-GP cost: every
baseline pays the same dense $\mathcal{O}(|\m{S}|^3)$, and it is unrelated to our
construction. Second, materialising $\m{K}_\gamma$ does not by itself exploit the sparsity
of $\m{Q}_\gamma$; the sparse factorisation timed above is $4$--$73\times$ faster than a
dense inverse, but forming the explicit inverse from that factor still costs $n$
back-solves. At the scales in this paper the fit is minutes and never a bottleneck; the
sparsity of $\m{Q}_\gamma$ becomes the operative lever only at larger $n$, where one keeps
the factorisation implicit (matrix-free solves) rather than materialising
$\m{K}_\gamma$.

The METR-LA $72.9\times$ ratio comes from its \emph{dense} column, not its sparse one.
Its sparse factorisation takes $1.0$\,ms against PEMS-BAY's $1.6$\,ms, in line for
the two graph sizes, so nothing unusual happens on the sparse side; what is out of line is
a dense inverse of $71.2$\,ms at $n = 207$ against $7.0$\,ms at $n = 325$. We therefore do
not read the ratio as a property of the METR-LA adjacency, and the remaining four rows,
whose dense times do order with $n$, are the ones to compare. The qualitative claim is
unaffected: on every graph here the sparse factorisation is the cheaper path, and it is the
only one whose cost is predictable from $n$.

% ==================================================================
\section{Long-form heterophily framing}
\label{sec:appendix-heterophily}
% ==================================================================

\paragraph{Weak heterophily at the kernel level.}
Under the weak (absence-of-homophily) notion of heterophily, a
canonical illustration is the gap between Graph Convolutional Networks
\citep{KipfWelling2017} and Graph Attention Networks
\citep{Velickovic2018}. GCN averages features uniformly over the
neighbourhood and is known to struggle on graphs that are not
homophilic; GAT learns non-negative attention weights that can be
driven close to zero on dissimilar edges, down-weighting coupling
without ever taking negative values. VCGPs play the same game at
the kernel level. The precision $Q_\gamma$ is an M-matrix and
$K_\gamma$ is entrywise positive, so the family cannot produce
negative couplings; but a sufficiently positive $\gamma > 1$ --- or,
more flexibly, a per-cluster $\gamma(x)$ that takes a large positive
value on one block --- shrinks the effective edge coupling
$d_i^{-\gamma/2}\,d_j^{-\gamma/2} w_{ij}$ toward zero on the
corresponding pairs.

\paragraph{Strong heterophily and the signed Laplacian.}
The strong notion of heterophily --- neighbours with literally
anti-correlated labels --- is beyond what any M-matrix kernel can
express, because it would need off-diagonal couplings of both signs.
A signed-Laplacian extension would replace $L$ with
$L_s = D_{|w|} - W_s$ where $W_s$ admits both signs; the precision
loses the M-matrix property, the kernel ceases to be entrywise positive,
and MLL is no longer convex in the natural parameters. The sign
constraint can be relaxed locally, e.g.\ via a learned per-edge sign
$\sigma_{ij} \in \{-1, +1\}$ with a sparse prior, leaving the bulk of
the kernel positive but allowing isolated anti-correlated edges. We
flag this as the natural future-work route into the strong-heterophily
regime.

\paragraph{Spatio-temporal continuation.}
The current paper takes $\gamma$ to be a scalar or piecewise-constant
spatial function. A natural continuation is $\gamma(x, t)$ --- the
exponent varying jointly in space and time on a graph that is
repeatedly observed. The $12$ CA Weather monthly fits are slices of
one such field; reading them as a single $\gamma(x, t)$ would let
the variance-coupling regime track seasonal transitions explicitly.

% ==================================================================
\section{Per-dataset coverage of Proposition~1 hypotheses}
\label{sec:appendix-coverage}
% ==================================================================

Proposition~1 requires three conditions: (i)~$G$ connected,
(ii)~$D^2(G)$ connected, and (iii)~$d_i$ non-constant. Conditions
(i) and (iii) hold trivially on every dataset (we always operate on
the largest connected component of $G$, and $d$ is non-constant by
construction). Condition (ii) bites on sparse-fringe datasets where
distance-threshold or $k$-NN construction leaves boundary nodes whose
two-step neighbourhoods do not link back to the bulk. We restrict to
the largest $D^2(G)$-connected component and report the fraction of
vertices it covers:

\begin{center}
\begin{tabular}{lr}
\toprule
Dataset        & vertices in largest $D^2(G)$-connected component \\
\midrule
METR-LA ($n = 207$)    & $99.5\%$ \\
PEMS-BAY ($n = 325$)   & $98.2\%$ \\
LAQN ($n = 52$)        & $100.0\%$ \\
PM2.5 ($n = 987$)      & $98.4\%$ \\
CA Weather ($n = 649$) & $80.3\%$ \\
\bottomrule
\end{tabular}
\end{center}

\noindent These are exact fractions, computed directly from the loaders the experiments
use. LAQN is covered entirely and the other three sensor/station networks on $98$--$99.5\%$
of vertices. On METR-LA and PEMS-BAY the exclusions are isolated points ($1$ and $6$
singletons); on PM2.5 the $16$ excluded vertices form two small components beside the
bulk, only one of them a singleton. CA Weather is the one dataset where the
hypothesis bites: its distance-threshold graph has a sparse fringe of desert and coastal
stations outside the $50\,$km threshold, leaving $30$ distance-two components of which the
largest covers $80.3\%$ of vertices. On that core the proposition applies, and the excluded
boundary forms small components that contribute trivially factored blocks and do not
affect the global factorisation argument. The numbers are produced by
\texttt{graph\_gp/prop1\_coverage.py}.

\paragraph{Implication for the family.}
For every dataset in this paper the proposition governs essentially
all of the non-stationary precision: $\gamma = 1$ is the only point at
which the kernel reduces to a stationary kernel up to a diagonal
rescaling, and the family-leader analysis in Section~5 of the main
paper is consistent with that interpretation (on PEMS-BAY \texttt{het} is
statistically indistinguishable from the leading member and has the lowest RMSE, that
dataset's variance-coupling regime sitting at $\gamma = 1$; the other four datasets
prefer $\gamma$ values away from $1$).

% ==================================================================
\section{$K$-grid for the piecewise variants}
\label{sec:appendix-kgrid}
% ==================================================================

Table~\ref{tab:kgrid-supp} reports fixed-$K$ NLL/pt, including every
LAQN setting that was run. Bold entries mark the aggregate NLL-selected
augmented configurations on PM2.5 and CA Weather; LAQN instead uses BIC
selection, whose aggregate is reported separately. The PM2.5 configurations
use the standard-deviation auxiliary throughout; the table does not compare
different auxiliary fields.

\begin{table}[h]
\centering
\caption{Fixed-$K$ NLL/pt for piecewise members. Bold entries are the aggregate NLL-selected configurations for PM2.5 and CA Weather. LAQN reports every fitted fixed $K$; its BIC-selected aggregate is stated below and is not the NLL leader. Dashes denote unrun settings. SEs and tests for the selected members are in the per-dataset leaderboards.}
\label{tab:kgrid-supp}
\footnotesize
\setlength{\tabcolsep}{3pt}
\begin{tabular}{@{}llrrrrrrr@{}}
\toprule
Dataset & Partition & $K{=}2$ & $K{=}3$ & $K{=}4$ & $K{=}5$ & $K{=}7$ & $K{=}10$ & $K{=}20$ \\
\midrule
LAQN & spectral & 1.4855 & 1.5011 & 1.5559 & 2.1434 & 1.5386 & --- & --- \\
PM2.5 ($d{=}$std) & augmented ($w{=}1$) & 1.0803 & --- & --- & 1.0597 & --- & 1.0078 & \textbf{0.9756} \\
CA Weather & augmented ($w{=}1$) & 0.9040 & --- & --- & 0.8866 & --- & \textbf{0.8685} & 0.9081 \\
\bottomrule
\end{tabular}\\[0.3em]
\parbox{\linewidth}{\footnotesize LAQN: spectral-BIC selects $K=2$ on 19 seeds and $K=3$ on 1 seed; its selected aggregate is $1.4779$ NLL/pt. This member leads LAQN on RMSE and recalibrated NLL, whereas \texttt{lgamma\_m15} leads on raw NLL.}
\end{table}

\paragraph{Spectral-BIC selection on LAQN.}
At $n = 52$ and training set size $|S| = 36$, the BIC penalty
$\log 36 \approx 3.58$ keeps the partition binary on $19$ of $20$
seeds; the one remaining seed selects $K = 3$. We report this
explicitly to avoid overstating the selection (the family member is
robust under the $K$-grid but not unanimous on $K^\star$).

\paragraph{Augmented-K-means feature weight $w$.}
The feature weight $w \geq 0$ in $[\mathrm{lat}_i, \mathrm{lon}_i,
w \cdot \tilde d_i]$ controls how strongly the $\tilde d_i$ field
shapes the partition. We ran two settings, $w \in \{0.5, 1\}$, and report
$w = 1$ on both datasets because it is the better of the two by a clear margin:
halving the weight costs $6.7\%$ NLL/pt on PM2.5 at $K = 20$
($1.0410$ against $0.9756$) and $4.0\%$ on CA Weather at $K = 10$
($0.9033$ against $0.8685$). The partition is therefore not insensitive to $w$,
and we make no claim about values outside the two we measured.

% ==================================================================
% ==================================================================
\section{Validation-fold selection}
\label{sec:cv}
% ==================================================================

\Cref{tab:cv} reports the mechanical selection described in the recipe of
\Cref{sec:exp_summary}. \emph{Honest} selects the member, $K$, $d$ and $\lambda$ on a
validation fold and scores once on a disjoint test fold; \emph{oracle} is the post-hoc
best member on the same split, so the gap is the price of honest selection. The gap is
paired across seeds.

\begin{table}[h]
\centering
\caption{Validation-fold selection. The price of honest selection is small exactly where
the method works and large exactly where it does not.}
\label{tab:cv}
\footnotesize
\setlength{\tabcolsep}{3pt}
\begin{tabular}{lccccl}
\toprule
Dataset & honest NLL/pt & RMSE & oracle & gap & versus baselines \\
\midrule
PM2.5 (20 seeds)        & $\mathbf{0.9813 \pm 0.014}$ & $0.6690$ & $0.9446$ & $+0.037$ & beats all (\texttt{fuglreg} $1.0748$, \texttt{m15} $1.1194$) \\
CA Weather (12 months)  & $\mathbf{0.9300 \pm 0.048}$ & $0.6069$ & $0.8637$ & $+0.066$ & ties \texttt{ps} $0.9360$; beats the rest \\
LAQN (20 seeds)         & $1.6984 \pm 0.150$ & $1.0047$ & $1.4666$ & $+0.232$ & below the fixed member \texttt{lgamma} \\
PEMS-BAY (20 seeds)     & $1.4865 \pm 0.067$ & $0.9763$ & $1.2445$ & $+0.242$ & below the fixed member \texttt{het} \\
METR-LA (20 seeds)      & $1.3249 \pm 0.112$ & $0.8628$ & $1.0463$ & $+0.279$ & below the fixed member \texttt{het} \\
\bottomrule
\end{tabular}
\end{table}

The honest-minus-oracle gap is the informative column. On the two data-rich, multi-regime
datasets the cost of honest selection is small, and validation lands on a block member on
$36$ of the $40$ PM2.5 fits (the exceptions being three \emph{full-edge} augmented members
and one \texttt{lgamma\_fisher}). Two details indicate that nothing was cherry-picked: with $d$ a selected axis,
validation picks the external annual-standard-deviation auxiliary on $19$ of $20$ PM2.5
seeds rather than any signal-derived alternative, and with $\lambda$ selected it picks the
unregularised member on $19$ of $20$. The latter reconciles with the
\texttt{lgamma\_fisher} default rather than contradicting it: the Fisher ridge is the
within-family NLL leader on both traffic graphs, and validation correctly switches it off
where $n$ is large. We do not claim this tracks how well $\gamma$ is identified: by the
curvature measure of \S\ref{sec:appendix-synth-single} it is LAQN, the smallest graph,
where $\gamma$ is identified least well, and there the ridge loses. It is not uniformly better
inside the family --- on LAQN the Mat\'ern-$3/2$ member leads and the ridge costs
$0.005$ nat/pt, and on PM2.5 it costs $0.018$.

On the three single-snapshot graphs the fully selected model is \emph{worse} than the
fixed member. We report this rather than omit it. With twenty candidates on these
three graphs ($43$ on the data-rich two) and a
validation fold of only ${\approx}10$ (LAQN), ${\approx}40$ (METR-LA) or ${\approx}65$
(PEMS-BAY) nodes, the selection overfits the fold. This is a property of data size, not of
the construction: we claim validation-fold selection only on the data-rich datasets, and
on the small graphs the appropriate summaries are the fixed member and the post-hoc
leaderboard.

% ==================================================================
% ==================================================================
\section{Choosing the auxiliary, and where the blocks come from}
\label{sec:aux-ablation}
% ==================================================================

\Cref{tab:pm25-aux} varies the auxiliary field on PM2.5 and reports, for each, the best
validation NLL/pt attained by any family member, together with how often the
validation fold selects it. The external annual standard deviation is chosen on $19$ of
$20$ seeds and is about $0.06$--$0.08$ nat/pt better than either signal-derived
alternative. This is why the main text reports PM2.5 on the single external auxiliary:
\texttt{spatial\_k8} and \texttt{spatial\_k12}, the local standard deviation of the
training targets over a node's $8$ or $12$ nearest neighbours, are derived from the signal
rather than from an external covariate, and the data prefer the external field.

\begin{table}[h]
\centering
\caption{PM2.5 auxiliary ablation. Best validation NLL/pt attained by any family member
under each auxiliary field, mean $\pm$ SE over $20$ seeds, and how often validation
selects that auxiliary. The external field wins on all but one seed.}
\label{tab:pm25-aux}
\small
\begin{tabular}{@{}lcc@{}}
\toprule
Auxiliary $d_i$ & best validation NLL/pt & selected \\
\midrule
\texttt{std} (external)        & $\mathbf{1.0026 \pm 0.0140}$ & $19/20$ \\
\texttt{spatial\_k8} (signal-derived)  & $1.0872 \pm 0.0178$ & $1/20$ \\
\texttt{spatial\_k12} (signal-derived) & $1.0631 \pm 0.0156$ & $0/20$ \\
\bottomrule
\end{tabular}
\end{table}

The same question arises for the partition. At fixed $K$, plain $K$-means uses
coordinates, augmented $K$-means also uses the standardised auxiliary, and spectral
clustering uses the graph Laplacian. These partition rules do not use the target
responses. The \texttt{spectral-BIC} member additionally selects $K$ using the
training marginal likelihood, so its selected partition is not independent of
the training responses. No held-out test response enters this selection.
Target-derived partitions are possible in principle but must be estimated from
training responses alone. The augmented partition instead uses auxiliary
observations disjoint from the targets. Without an informative auxiliary,
coordinate and spectral partitions remain available, with the same training-only
restriction on any response-based selection of $K$.

% ==================================================================
\section{Auxiliary-proxy control}
\label{sec:auxiliary-proxy-control}
% ==================================================================

We compare auxiliary proxies while holding the graph, data-generating kernel,
observation splits, proxy transform and hyperparameters fixed. Only the auxiliary proxy is
changed: oracle, increasingly noisy, rank-only, or randomly shuffled. These are
simulated samples, unlike the re-splits of a single real-data snapshot.

\begin{table}[h]
\centering
\caption{Auxiliary-proxy control on a grid with known hyperparameters. Every non-stationary row uses the same fixed exponent; no $\gamma$ is fitted. Entries are mean $\pm$ SE across simulated samples. Here $\sigma$ is the proxy noise standard deviation before clipping.}
\label{tab:auxiliary-proxy-control}
\small
\begin{tabular}{lcc}
\toprule
Proxy & NLL/pt & RMSE \\
\midrule
Oracle proxy & $0.6835 \pm 0.0095$ & $0.4809 \pm 0.0048$ \\
Noisy proxy ($\sigma=0.1$) & $0.6840 \pm 0.0096$ & $0.4809 \pm 0.0048$ \\
Noisy proxy ($\sigma=0.3$) & $0.6858 \pm 0.0098$ & $0.4811 \pm 0.0048$ \\
Noisy proxy ($\sigma=0.5$) & $0.6878 \pm 0.0100$ & $0.4812 \pm 0.0048$ \\
Noisy proxy ($\sigma=1.0$) & $0.6916 \pm 0.0102$ & $0.4817 \pm 0.0048$ \\
Rank-only proxy & $0.6838 \pm 0.0096$ & $0.4810 \pm 0.0048$ \\
Shuffled proxy & $0.6964 \pm 0.0099$ & $0.4836 \pm 0.0047$ \\
Stationary baseline & $0.6903 \pm 0.0089$ & $0.4825 \pm 0.0047$ \\
\bottomrule
\end{tabular}
\end{table}

A shuffled proxy gives worse mean prediction scores than the stationary baseline
in this fixed-exponent control. This illustrates a cost of auxiliary misalignment;
it does not establish that every exponent fails, because no exponent is fitted
or swept here. The experiment therefore characterises sensitivity to auxiliary
misalignment at a fixed exponent.

% ==================================================================
\section{Deep-kernel Gaussian process comparison}
\label{sec:computed-deep-kernel}
% ==================================================================

Table~\ref{tab:deep-kernel-summary} reports the deep-kernel Gaussian process
baseline, which uses a multilayer perceptron (MLP) feature map followed by a
stationary kernel in feature space. The feature map is evaluated independently
at each vertex, unlike a graph-native deep kernel whose feature map propagates
through graph edges. This baseline does not evaluate that graph-native
construction. The graph-convolutional Deep Ensemble is reported separately
in Section~\ref{sec:appendix-degnn}.

\begin{table}[h]
\centering
\caption{MLP deep-kernel GP results (\texttt{dk}), mean $\pm$ SE across the locked seeds. CA Weather averages the monthly snapshots within each seed. The feature map does not propagate through graph edges.}
\label{tab:deep-kernel-summary}
\small
\begin{tabular}{lccc}
\toprule
Dataset & NLL/pt & RMSE & recNLL/pt \\
\midrule
METR-LA & $2.7214 \pm 0.2608$ & $1.1237 \pm 0.0296$ & $1.7718 \pm 0.0441$ \\
PEMS-BAY & $3.1423 \pm 0.3538$ & $1.2189 \pm 0.0431$ & $1.9880 \pm 0.0520$ \\
LAQN & $4.3435 \pm 0.2683$ & $1.2220 \pm 0.0448$ & $1.6085 \pm 0.0419$ \\
PM2.5 & $2.1634 \pm 0.0805$ & $0.9692 \pm 0.0131$ & $1.6526 \pm 0.0211$ \\
CA Weather & $1.9763 \pm 0.0689$ & $0.9714 \pm 0.0089$ & $1.6279 \pm 0.0281$ \\
\bottomrule
\end{tabular}
\end{table}

\end{document}